\documentclass[final]{cvpr}

\usepackage{times}

\usepackage{graphicx}
\usepackage{float}
\usepackage{subfig}
\usepackage{multirow}
\usepackage{amsmath}
\usepackage{amssymb}
\usepackage{booktabs}
\usepackage{algpseudocode}

\usepackage{bbding}
\usepackage{pifont}

\usepackage[ruled, lined, linesnumbered, commentsnumbered, longend]{algorithm2e}
\usepackage{xcolor}

\SetCommentSty{mycommfont}

\usepackage[pagebackref,breaklinks,colorlinks]{hyperref}

\usepackage[capitalize]{cleveref}
\crefname{section}{Sec.}{Secs.}
\Crefname{section}{Section}{Sections}
\Crefname{table}{Table}{Tables}
\crefname{table}{Tab.}{Tabs.}

\def\cvprPaperID{7149} 
\def\confName{CVPR}
\def\confYear{2023}

\begin{document}

\title{4D-Editor: Interactive Object-level Editing \\ in Dynamic Neural Radiance Fields via Semantic Distillation}

\author{Dadong Jiang$^{\ast}$ \and Zhihui Ke$^{\ast}$ \and Xiaobo Zhou$^{\dagger}$ \and Tie Qiu \and Xidong Shi
\\
College of Intelligence and Computing, Tianjin University\\
{\tt\small \{patrickdd, kezhihui, xiaobo.zhou, qiutie, suif\}@tju.edu.cn}}


\twocolumn[{
\maketitle
\begin{figure}[H]
\hsize=\textwidth 
\centering
\vspace{-15mm}
\includegraphics[width=18cm, trim=220 250 300 30, clip]
{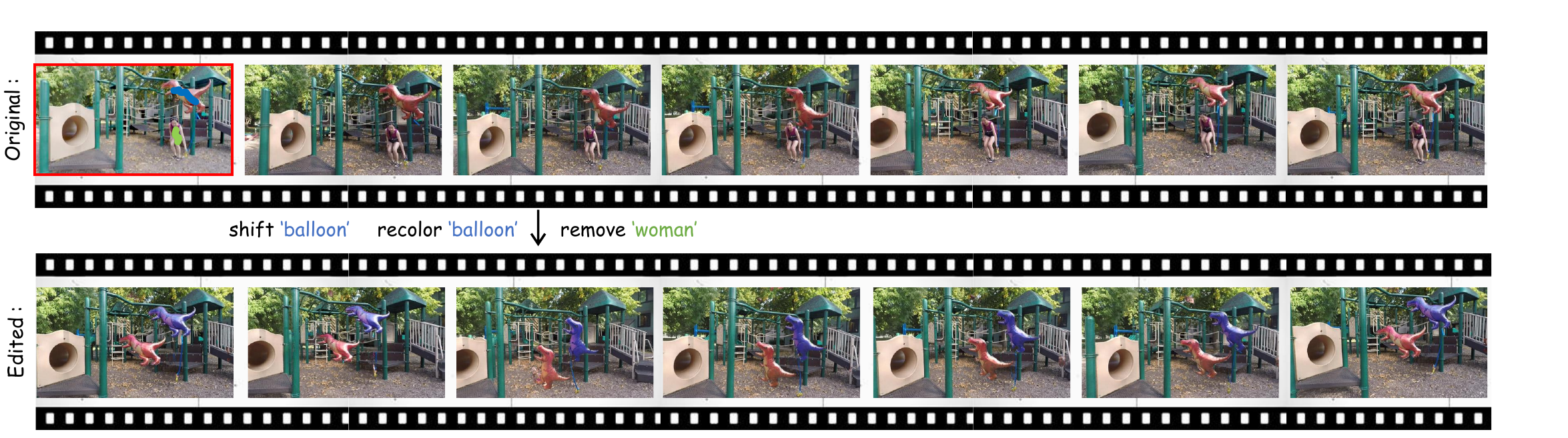}
\caption{
4D-Editor can interactively edit objects in a dynamic NeRF. For example, with strokes drawn on a reference frame by users, 4D-Editor can remove the human, recolor and shift the balloon within the dynamic NeRF.
After the novel view synthesis, the women disappears, the balloon is changed into purple and shifted in both spatial and temporal dimensions.
}
\label{fig:head}
\end{figure}
}]
    

\begin{abstract}
This paper targets interactive object-level editing (\textit{e.g.}, deletion, recoloring, transformation, composition) in dynamic scenes.
Recently, some methods aiming for flexible editing static scenes represented by neural radiance field (NeRF) have shown impressive synthesis quality, while similar capabilities in time-variant dynamic scenes remain limited.
To solve this problem, we propose 4D-Editor, an interactive semantic-driven editing framework, allowing editing multiple objects in a dynamic NeRF with user strokes on a single frame.
We propose an extension to the original dynamic NeRF by incorporating a hybrid semantic feature distillation to maintain spatial-temporal consistency after editing.
In addition, we design Recursive Selection Refinement that significantly boosts object segmentation accuracy within a dynamic NeRF to aid the editing process.
Moreover, we develop Multi-view Reprojection Inpainting to fill holes caused by incomplete scene capture after editing.
Extensive experiments and editing examples on real-world demonstrate that 4D-Editor achieves photo-realistic editing on dynamic NeRFs.
Project page: \href{https://patrickddj.github.io/4D-Editor}{https://patrickddj.github.io/4D-Editor}
\end{abstract}
\vspace{-4mm}

\let\thefootnote\relax\footnotetext{$\ast$ Equal Contribution}
\let\thefootnote\relax\footnotetext{$\dagger$ Corresponding Author}

\section{Introduction}
\label{sec:intro} 
With the popularity of Neural Radiance Field (NeRF~\cite{mildenhall2020nerf}) and its extensions~\cite{fridovich2022plenoxels, M_ller_2022, chen2022tensorf, tretschk2021non, nerfplayer, gao2021dynamic, liu2023robust, jang2022dtensorf, park2021hypernerf} enabling observation on dynamic real-world scenes in a free-viewpoint way, there is a critical demand for editing NeRF to support downstream applications, \textit{e.g.}, VR/AR, computer animation, and education.
Semantic-NeRF~\cite{zhi2021place} utilizes existing semantic labels for scene understanding and facilitates object-level editingg, while it requires mutual labels.
Therefore, some recent studies~\cite{kobayashi2022decomposing, tschernezki2022neural, goel2023interactive} adopt semantic distillation to extract 3D semantic features in a self-supervised way from large pre-trained models like DINO~\cite{caron2021emerging} or LSeg~\cite{li2022languagedriven}, which can generate open-vocabulary scene semantic labels as prior information.
Despite these methods can achieve fine editing results in 3D static NeRFs, they are constrained to edit one 4D dynamic NeRF due to its spatial-temporal variant.

Our goal is to interactively edit target objects within a dynamic NeRF. The closest work to ours is NeuPhysics~\cite{qiao2022NeuPhysics}, which allows partially editing dynamic scenes.  
It simply decomposes a scene into the dynamic foreground and the static background, which means that the foreground and background are treated as distinct entities, allowing only the complete editing of either the foreground or background.
Thus, editing a single object within a scene containing multiple objects is not supported.(\textit{e.g.}, recolor a street sign in a static background or remove one specific car when existing multiple moving cars).
Moreover, NeuPhysics relies on a mesh proxy for editing and does not provide any editing interface, which is user-friendliness. The differences between our work and existing works are listed in Table~\ref{table:comparison}.


In this paper, we propose an interactive object-level editing framework for dynamic NeRFs, named 4D-Editor, which enables users to select and edit different objects by strokes on one reference frame, the modification effect of which will propagate throughout the entire dynamic NeRF.
However, maintaining both multi-view consistency and time-variant consistency becomes challenging when applying modifications on an observation view to the whole time series.
Therefore, we employs Hybrid Semantic Feature Distillation to extract and distill semantic information from a pre-trained DINO\cite{caron2021emerging} into hybrid semantic radiance field, which contains 3D and 4D semantic features separately to aid object segmentation and editing process.
Nevertheless, it is difficult to precisely segment edited objects in dynamic NeRFs due to few user strokes and fill \textit{holes} with spatial-temporal consistency after editing owning to limited observation views. 
To solve above problems, we propose Recursive Selection Refinement method to ensure accurate matching of all target objects while preserving unrelated areas. 
Furthermore, we propose a Multi-view Reprojection Inpainting strategy to fill \textit{holes} caused by insufficient scene capture, which involves completing visible parts through observations from multiple perspectives and generating invisible parts using an inpainting model. 
The editing process takes only about 1-2 seconds to generate a novel modified view.

\begin{table}
\centering
\setlength{\tabcolsep}{1.2mm}{
\begin{tabular}{c | c c c c}
Method & Dynamic & Interactive &  Object-Level \\ 
\hline
Ours     & \checkmark & \checkmark & \checkmark \\
NeuPhysics\cite{qiao2022NeuPhysics }     & \checkmark & \ding{55} & \ding{55} \\ 
N3F\cite{tschernezki2022neural}     & \checkmark\kern-1.2ex\raisebox{1ex}{\rotatebox[origin=c]{125}{\textbf{--}}} & \ding{55} & \checkmark\kern-1.2ex\raisebox{1ex}{\rotatebox[origin=c]{125}{\textbf{--}}} \\ 
ISRF\cite{goel2023interactive}     & \ding{55} & \checkmark & \checkmark \\ 
\end{tabular}
}
\vspace{2mm}
\caption{Comparison with prior methods.}
\label{table:comparison}
\vspace{-4mm}
\end{table}

We summarize our contributions as follows:
\begin{itemize}
    \item We propose 4D-Editor, to our knowledge, the first interactive editing framework which is used to edit multiple objects within dynamic NeRFs with user strokes on 2D images, delivering spatial-temporal consistency throughout the dynamic scene. 4D-Editor enables diverse editing operations \textit{e.g.}, deletion, recoloring, transformation, composition.
    
    \item We incorporate a Hybrid Semantic Feature Distillation to extract semantic information in 4D space from pre-trained DINO models for maintaining spatial-temporal consistency after editing operations.
    
    \item We propose Recursive Selection Refinement, an accurate 4D object segmentation method that enables rapid and precise selection of target objects in the dynamic NeRF in a recursive manner. Our experimental results confirm its accuracy and efficiency.
    
    \item Multi-view Reprojection Inpainting is developed to fill holes  caused by incomplete scene capture from sparse views, particularly after removal operation.
\end{itemize}


\begin{figure*}[tbp]
    \centering
    \includegraphics[width=18cm, trim=530 160 280 50, clip]{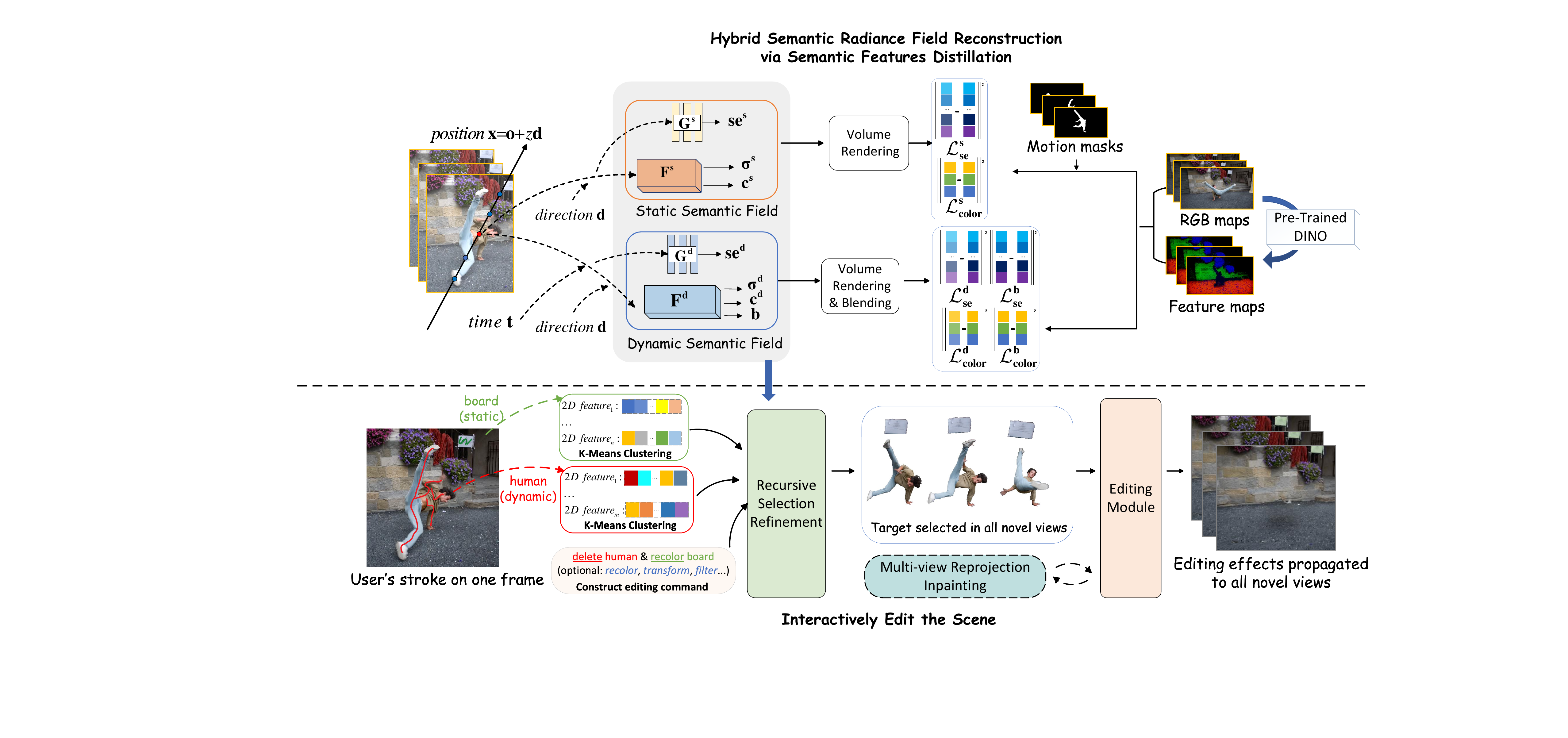}
    \caption{\textbf{4D-Editor framework overview.} 
    The dynamic scene is represented by hybrid radiance fields $F^s$, $F^d$ and corresponding semantic fields $G^s$ and $G^d$ which are distilled from DINO teacher model.
    A user can choose a reference view, mark on desired edited objects and assign desirable operations. For example, \textbf{delete} dynamic human(red[\textcolor{red}{$\sim$}] stroke) and \textbf{recolor} static board(green[\textcolor{green}{$\sim$}] stroke).
    Then, two groups of 2D semantic features are collected and clustered by K-Means to recursively match the corresponding 4D features in dynamic NeRF to achieve precise object segmentation.
    Finally, desired editing operations on these objects are applied and the effect of editing is spread to the whole NeRF(the human disappeared and the board turned green).
    }
    \label{fig:system_framework}
    \vspace{-4mm}
\end{figure*}

\section{Related work}
\label{sec:releated_work}

\noindent \textbf{Dynamic Scene Representations.} 
In the past few years, dynamic scene representations~\cite{li2022neural, yoon2020novel, jang2022dtensorf, kplanes_2023, liu2023robust, wang2023mixed, gan2023v4d, yang2023real, lewin2023dynamic, liu2022devrf, wang2023neural, cao2023hexplane, tretschk2021non} have experienced great development both in terms of reconstruction quality and training speed. 
DyNeRF~\cite{li2022neural} is an early contribution to this field, introducing expressive time-variant latent codes into implicit volume representations for reconstructing dynamic scenes. However, the training speed of DyNeRF is extremely slow, requiring 7 days.
Thus, recent explicit volume representation methods like K-Planes\cite{kplanes_2023} and D-TensoRF~\cite{jang2022dtensorf} use multiple low-dimension planes or vectors to represent the 4D dynamic scene, thereby greatly accelerating training and rendering speed.
Different from directly learning dynamic scene representation, DynamicNeRF~\cite{yoon2020novel}, RobustNeRF~\cite{liu2023robust}, and MixVoxels~\cite{wang2023mixed} divide the scene into static and dynamic parts and utilize hybrid radiance field representation to model them separately.




\noindent \textbf{NeRF Editing.}
One direct approach to editing NeRF is to convert it into textured mesh~\cite{yang2022neumesh, yuan2022nerf, liu2023neural, yang2023learning}, enabling the use of existing 3D editing tools(\textit{e.g.}, Blender~\cite{blender}) for texture editing. 
However, this method poses challenges on baking large scenes with high quality and is limited to reconstructing static scenes. 
With the development of language-guided models, researchers utilize CLIP model~\cite{radford2021learning} or diffusion model~\cite{muller2023diffrf, metzer2023latent, instructnerf2023, zhang2023text2nerf, seo2023dittonerf, park2023ednerf} to perform NeRF editing using text prompts and image patches. 
Additionally, researchers have proposed blending editing methods~\cite{bao2023sine, gordon2023blended}, which combine an auxiliary editing field with original NeRF to support creative editing. Some methods employ inpainting models~\cite{mirzaei2023spin, yin2023ornerf, wang2023inpaintnerf360, mirzaei2023referenceguided, liu2022nerfin} to fill holes after object removal. Nevertheless, all above methods are limited to editing static NeRFs.
Regarding editing of dynamic NeRF, NeuPhysics~\cite{qiao2022NeuPhysics } utilizes time-invariant signed distance function(SDF) with a deformation field to reconstruct dynamic scenes, while not supports object-level editing. 
Recently, N3F~\cite{tschernezki2022neural} and ISRF~\cite{goel2023interactive} treat pre-trained semantic models as a teacher and distill 2D semantic features from a teacher model into feature fields, enabling object segmentation and editing in 3D space. However, N3F not only lacks interfaces for interactive editing but also exhibits poor editing performance in dynamic scenes due to imprecise object segmentation. In contrast, our proposed 4D-Editor enables interactive editing of multiple objects in a dynamic NeRF and maintains spatial-temporal consistency.

\noindent \textbf{NeRF with Semantics.} 
Recent methods introduce semantic features into NeRF for spatial semantic segmentation~\cite{kundu2022panoptic, fu2022panoptic, zhi2021place, vora2021nesf, chen2022sem2nerf}. 
Semantic-NeRF~\cite{zhi2021place} encodes scene semantic information with appearance and geometry, achieving interactive segmentation using semantic label propagation.
PNF~\cite{kundu2022panoptic} optimizes an object-aware neural scene representation that decomposes a scene into a set of objects and background using pseudo-supervision from predicted semantic segmentation.
However, these methods require manual annotated labels. In contrast to above supervision way, our method distill semantic features from a pre-trained semantic model into dynamic NeRFs in a self-supervision manner.

\section{Method}
\label{sec:method}
In this section, we first briefly introduce the background of hybrid radiance field representation of dynamic scenes (Sec.~\ref{sec:preliminary}).
Subsequently, we propose our Hybrid Semantic Features Distillation method which serves as a guidance for editing one dynamic NeRF (Sec.~\ref{sec:method_hybrid_distillation}).
We then detail the interactive editing pipeline including K-means clustering, 2D-4D feature matching, recursive selection refinement (Sec.~\ref{sec:method_recursive_refinement}) and editing module (Sec.~\ref{sec:method_editing}).
Finally, we demonstrate the Multi-view Reprojection Inpainting method for hole-filling in invisible areas after editing
(Sec.~\ref{sec:progressive_inpainting}).
The proposed framework is shown in Fig.~\ref{fig:system_framework}.

\subsection{Preliminary: Hybrid Radiance Field Reconstruction of Dynamic Scenes}
\label{sec:preliminary}
Hybrid radiance field representations of dynamic scenes, such as DynNeRF~\cite{gao2021dynamic} and RoDyn-NeRF~\cite{liu2023robust}, usually consist of a static radiance field $F^s$ and a dynamic radiance field $F^d$.
Given a 3D position $ \textbf{x} \in \mathbb{R}^3$ with its normalized viewing direction $\textbf{d} \in \mathbb{R}^3$ and time $t \in \mathbb{R}$, 
$F^s$ maps radiance values as time-invariant density $\sigma^s$ and RGB color $\textbf{c}^s$ for static background:
\begin{equation}
    (\sigma^s, \textbf{c}^s) = F^s(\textbf{x},\textbf{d})
\end{equation}

While $F^d$ maps radiance values as time-variant density $\sigma^d$ and RGB color $\textbf{c}^d$ for dynamic foreground. 
Furthermore, $F^d$ also predicts blending weight $b$ for blending the output of $F^s$ and $F^d$:
\begin{equation}
    (\sigma^d, \textbf{c}^d, b) = F^d(\textbf{x},\textbf{d},t)
\end{equation}

The calculated density and color are then used in volume rendering alone the ray $\textbf{r}$ emitted from the camera to obtain corresponding pixel color:
\begin{equation}
\begin{split}
    \hat{C}(\textbf{r}) = & \sum_{i=1}^{N} T(u_i) \cdot (\alpha(\sigma^s(u_i) \delta_i) \cdot \textbf{c}^s(u_i) \cdot b\\ 
    & + \alpha(\sigma^d(u_i) \delta_i) \cdot \textbf{c}^d(u_i) \cdot (1-b))
    \label{con:blending_color}
\end{split}
\end{equation}
\begin{equation}
    T(u_i) = exp\left(\sum_{j=1}^{i-1} \sigma^s(u_j)\delta_j b + \sigma^d(u_j)\delta_j (1-b)\right)
\end{equation}
where $\alpha(z) = 1-exp(-z)$, $\delta_i = u_{i+1} - u_i$ is the distance between two neighbor sampled points along the ray, the $N$ points $\{u_i\}_{i=1}^N$ are uniformly sampled between near plane and far plane~\cite{mildenhall2020nerf}, and $T(u_i)$ indicates the accumulated transmittance.

\subsection{Hybrid Semantic Features Distillation}
\label{sec:method_hybrid_distillation}

In order to maintain the spatial-temporal consistency during object-level editing, we use two semantic fields to store semantic features of static and dynamic parts of a scene, respectively, which are denoted as $G^s$ and $G^d$. 
We employ a large pre-trained teacher model (\textit{e.g.}, DINO\cite{caron2021emerging}) to distill semantic features into two semantic fields, which serves as guidance for editing.
Therefore, the static semantic field is represented by $F^s$ and $G^s$ that stores time-invariant semantic features. Similarity, the dynamic semantic field is represented by $F^d$ and $G^d$ that stores time-variant features as shown in Fig.~\ref{fig:system_framework}.

We obtain the time-invariant semantic feature $\textbf{se}^s \in \mathbb{R}^C$ and time-variant semantic feature $\textbf{se}^d \in \mathbb{R}^C$ of a sampled point according to follows:
\begin{equation}
    \textbf{se}^s = G^s(\textbf{x}), \quad
    \textbf{se}^d = G^d(\textbf{x}, t)
\end{equation}
Note that we disregard view direction $\textbf{d}$ due to the direction-agnostic nature of scene semantics.
Specifically, given a set of $N$ consecutive frames $\mathcal{I}: {\{I_i\}}_{i=1}^N \in \mathbb{R}^{H \times W \times 3}$, we utilize the DINO ViT-b8 model to generate corresponding semantic feature maps $\in \mathbb{R}^{H/8 \times W/8 \times C}$. Then, we upsample these low-resolution feature maps through an upsampling layer to output the final feature maps ${\{Se_i\}}_{i=1}^N \in \mathbb{R}^{H \times W \times C}$.
Next, we employ volume rendering to obtain the pixel-aligned semantic features $\hat{Se}^s(\textbf{r})$ and $\hat{Se}^d(\textbf{r})$, which represent the accumulated semantic feature along a ray $\textbf{r}$:
\begin{equation}
\begin{split}
    &\hat{Se}^s(\textbf{r}) = \sum_{i=1}^{N} T^s(u_i)\alpha(\sigma^s(u_i) \delta_i) \textbf{se}^s(u_i), \\
    &\hat{Se}^d(\textbf{r}) = \sum_{i=1}^{N} T^d(u_i)\alpha(\sigma^d(u_i) \delta_i) \textbf{se}^d(u_i)
\end{split}
\end{equation}
where $T^s(u_i)=exp(\sum_{j=1}^{i-1}\sigma^s(u_j)\delta_j)$ and $T^d(u_i) = exp(\sum_{j=1}^{i-1}\sigma^d(u_j)\delta_j)$. 

Finally, we calculate the final semantic feature by blending semantic features of $G^s$ and $G^d$ outputs, similar to Equation~\ref{con:blending_color}: 

\begin{equation}
\begin{split}
    \hat{Se}^{b}(\textbf{r}) = & \sum_{i=1}^{K} T^{b}(u_i)(\alpha(\sigma^s(u_i)\delta_i) \textbf{se}^s(u_i) b \\
    & + \alpha(\sigma^d(u_i) \delta_i) \textbf{se}^d(u_i) (1-b))
\end{split}
\end{equation}

We add three new losses to train $G^s$ and $G^d$: $\mathcal{L}_{se}^s$ for pixels belonging to the static part, $\mathcal{L}_{se}^d$ and $\mathcal{L}_{se}^b$ for all pixels. We treat the output of the teacher model as the ground truth. By minimizing the difference between predicted features and the ground truth, these two semantic fields can learn scene semantics.
\begin{equation}
\begin{split}
    & \mathcal{L}_{se}^s = \sum_{\textbf{r} \in \mathcal{R}^s} {\lVert Se(\textbf{r}) - \hat{Se}^s(\textbf{r}) \rVert}_2^2, \\
    & \mathcal{L}_{se}^d = \sum_{\textbf{r} \in \mathcal{R}^s + \mathcal{R}^d} {\lVert Se(\textbf{r}) - \hat{Se}^d(\textbf{r}) \rVert}_2^2, \\
    & \mathcal{L}_{se}^{b} = \sum_{\textbf{r} \in \mathcal{R}^s + \mathcal{R}^d} {\lVert Se(\textbf{r}) - \hat{Se}^b(\textbf{r}) \rVert}_2^2
\end{split}
\end{equation}
where $\mathcal{R}^s$, $\mathcal{R}^d$ are sampled rays from static and dynamic part of a scene, respectively.
The total semantic loss is
\begin{equation}
    \mathcal{L}_{se} = \mathcal{L}_{se}^s + \lambda_1 \mathcal{L}_{se}^d + \lambda_2 \mathcal{L}_{se}^{b}
\end{equation}


\subsection{Recursive Selection Refinement}
\label{sec:method_recursive_refinement}
\noindent \textbf{K-Means Query for Multi-Objects Selection.}
In our 4D-Editor framework, users can mark desired objects with a brush on a reference frame.
Based on user's strokes, 4D-Editor extracts target 2D semantic features from corresponding feature maps generated by DINO.
These semantic features are utilized to construct different queries for matching multiple objects in semantic fields  $G^s$ and $G^d$. Then, we can use these queries to match 2D-4D features to segment target objects in hybrid semantic radiance field. 

\begin{figure}[tbp]
\vspace{-3mm}
\centering
    \subfloat[Average Feature]{
        \includegraphics[width=0.14\textwidth]{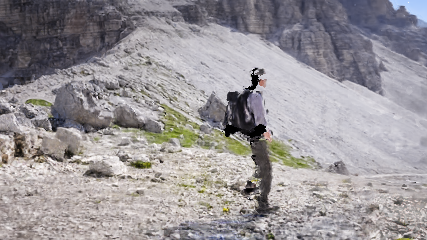}
        \label{fig:ablation_average_feature}
    }
    \hfill
    \subfloat[K-Means Only]{
        \includegraphics[width=0.14\textwidth]{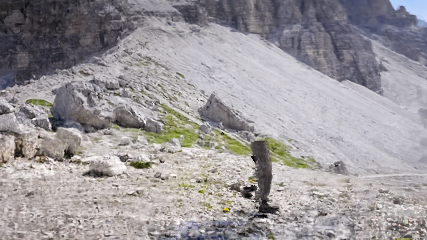}
		\label{fig:ablation_depth_1}
    }
    \hfill	
    \subfloat[$K=3$]{
        \includegraphics[width=0.14\textwidth]{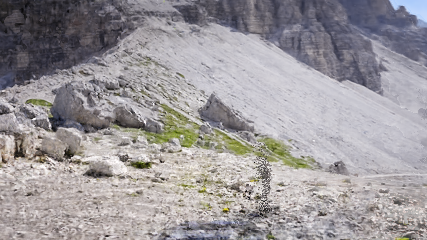}
		\label{fig:ablation_depth_3}
    }
    \hfill

    \vspace{-3mm}
    
    \subfloat[$K=5$]{
        \includegraphics[width=0.14\textwidth]{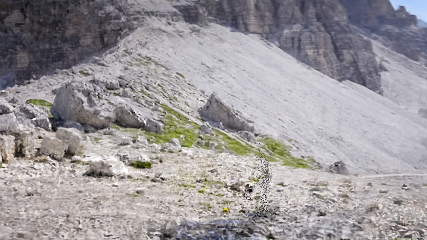}
		\label{fig:ablation_depth_5}
    }
    \hfill	
    \subfloat[$K=10$]{
        \includegraphics[width=0.14\textwidth]{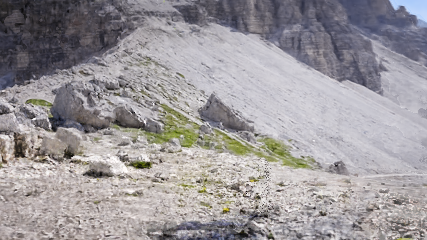}
		\label{fig:ablation_depth_10}
    }
    \hfill	
    \subfloat[$K=30$]{
        \includegraphics[width=0.14\textwidth]{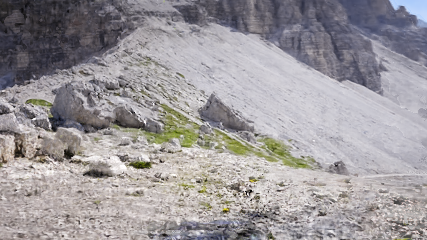}
		\label{fig:ablation_depth_30}
    }
    \vspace{2mm}
\caption{
    \textbf{Refinement with different recursive number.}
    Compared to \textbf{(a)} and \textbf{(b)}, our method \textbf{(c)-(f)} achieve great improvement in object removal.
    With larger recursion number, artifacts are eliminated, while maintaining spatial-temporal consistency in other areas. Similar to the convergence of ray-tracing, our method also achieve high quality when $K=10$ ($\alpha=0.6, \beta=0.1$). The benefits generated by a larger $K$ are very small.
}
\label{fig:ablation_refinement}
\vspace{-3mm}
\end{figure}

However, the strokes provided by the user are sparse, resulting in naturally insufficient and inexpressive semantic features, where a simple query such as averaging features (Fig.~\ref{fig:ablation_average_feature}), can lead to incorrect 2D-4D feature matches.
Therefore, inspired by ISRF\cite{goel2023interactive},
we utilize K-Means to group the most significant and relevant features for each individual object, aiming to enhance the accuracy of feature matching.

\noindent \textbf{Recursive Refinement.} 
Despite employing K-Means for effective 2D-4D feature matching, achieving accurate object segmentation still remains challenging:
Owing to 8x up-sampled feature maps, unsupervised semantic feature distillation inherently leads to imprecise semantic segmentation in hybrid semantic radiance field, especially for the confusion between object edges and background (Fig.~\ref{fig:ablation_depth_1}).

Given a $M$ K-means clustered query $\gamma \in \mathbb{R}^{M \times C}$, a set of sampled points $\mathcal{U}$, and corresponding indexes $\mathcal{A}$, we can obtain the feature distance $d(\gamma, u_i)$ between one sampled point $u_i \in \mathcal{U}$ and query $\gamma$ according to $d(\gamma, u_i) = \min_M({\lVert \gamma - se(u_i) \rVert}_2)$. 
However, it is difficult to directly set the threshold $\alpha$ of feature distance to select potential points where $d(\gamma, u_i) < \alpha$, due to the semantic ambiguity between object edges and background mentioned above. Here, a slightly higher threshold may result in selection of unexpected areas, while a lower threshold cannot ensure that the entire object is selected, leading to obvious ``artifacts'' in rendered views.
Another native solution is to use neighbour points(\textit{e.g.}, within a unit sphere) to indirectly judge the validity of the current point.
However, the radius is still a threshold that is infeasible to set due to the ambiguousness of object edges.

To solve this problem, we propose recursive selection refinement(RSR) algorithm that is inspired by the ray-tracing process~\cite{shirley2003realistic}, which estimate unbiased indirect illumination through recursively computing the intersection of rays and surfaces.
RSR algorithm recursively refine the object selection instead of relying on a fixed threshold. Thus, we not only avoid heavy manual threshold setting but also achieve precise object selection, especially object edges. Specifically, we add exploration range $\beta$ and divide all sampled points into three sets: 
(1) Valid point set $\mathcal{V}$, the points in which comprise the main part of the object. 
(2) Possible point set $\mathcal{P}$, which contains points are likely to be on the object surface. 
(3) Impossible point set $\mathcal{Q}$, which contains points far away from the object.
Table~\ref{table:three_types} shows the differences. Note that our proposed RSR algorithm is not sensitive to parameter $\beta$ (Refer to \textcolor{blue}{supplementary material Sec. E} for details).

\begin{table}[tbp]
\vspace{-3mm}
\centering
\setlength{\tabcolsep}{1.2mm}{
\begin{tabular}{c  c}
\hline
$u_i \in \mathcal{V}$     & $ d(\gamma, u_i) \leq \alpha $ \\
$u_i \in \mathcal{P}$     & $ \alpha < d(\gamma, u_i) \leq \alpha + \beta $ \\
$u_i \in \mathcal{Q}$     & $ d(\gamma, u_i) > \alpha + \beta $ \\
\hline
\end{tabular}
}
\vspace{2mm}
\caption{Ranges of feature distance among $\mathcal{V}, \mathcal{P}, \mathcal{Q}$.}
\label{table:three_types}
\vspace{-4mm}
\end{table}

Then, we need to distinguish these sampled points truly belonging to the target object in possible point set $\mathcal{P}$. 
We record their original indexes and apply random offsets to points in $\mathcal{P}$ to nudge them into valid point set and impossible point set, thereby forming new valid point set $\mathcal{V}^{\prime}$, possible point set $\mathcal{P}^{\prime}$, and impossible point set $\mathcal{Q}^{\prime}$ as shown in Algorithm~\ref{alg:ReSeR}.
We repeat this recursive refinement process until reaching the maximum recursion number $K$ or the possible point set is empty. 
Similar to ray-tracing, we also can obtain the approximate unbiased estimation on the object segmentation (or object selection) as demonstrated in Fig.~\ref{fig:ablation_refinement} and Table~\ref{table:points_in_each_iteartion}.

\begin{algorithm}[tbp]
\SetKwInput{KwInput}{Input}                
\SetKwInput{KwOutput}{Output}              
  
  \KwInput{$\mathcal{U}, \mathcal{A}, K, \alpha, \beta, \gamma, s, k=0$}
  \KwOutput{valid point index set $\mathcal{W}$}

  \SetKwFunction{FRSE}{RSR}
  \SetKwFunction{FOFFSET}{offset}
  \SetKwFunction{FRAND}{randn}
  \SetKwFunction{FAPPEND}{append}
 
  \SetKwProg{Fn}{Def}{:}{}
  \Fn{\FRSE{$\mathcal{U}, \mathcal{A}, k, \alpha, \beta, \gamma, s$}}{

    $\mathcal{W} \leftarrow [ \quad ] $
    
    \textbf{if} {$k = K$ or $\mathcal{U} = \varnothing$}: \KwRet $\mathcal{W}$ 

    \tcp{Calculate feature distances}
    $\mathcal{D} \leftarrow \{d(\gamma, u_i), u_i \in \mathcal{U}\} $

    \tcp{Store valid points $\&$ indexes}
    $\mathcal{V} \leftarrow \{d(\gamma, u_i) \leq \alpha, d(\gamma, u_i) \in \mathcal{D}\}$
    
    $\mathcal{W}.\FAPPEND$(index of $\mathcal{V}$ from $\mathcal{A}$)

    \tcp{Dismiss impossible points}
    $\mathcal{Q} \leftarrow \{ d(\gamma, u_i) > \alpha + \beta, d(\gamma, u_i) \in \mathcal{D}\}$

    \tcp{Get possible points and indexes}
    $\mathcal{P} \leftarrow \{\alpha < d(\gamma, u_i) \leq \alpha + \beta, d(\gamma, u_i) \in \mathcal{D}\} $
    
    $\mathcal{A^{\prime}} \leftarrow$ index of $\mathcal{P}$ from $\mathcal{A}$
    
    \tcp{Apply random offsets}

    $\mathcal{P^{\prime}} \leftarrow \FOFFSET(\mathcal{P}, \FRAND(0, s)) $

    $\mathcal{V^{\prime}} \leftarrow$ \FRSE($\mathcal{P^{\prime}}, \mathcal{A^{\prime}}, k+1, \alpha, \beta, \gamma, s)$

    $\mathcal{W}.\FAPPEND$(index of $\mathcal{V^{\prime}}$ from $\mathcal{A}$)
    
    \KwRet $\mathcal{W}$
    }
\caption{Recursive Selection Refinement}
\label{alg:ReSeR}
\end{algorithm}

\subsection{Editing Module}
\label{sec:method_editing}
After the object segmentation, for each editing object, we can obtain two sampling point sets: $\mathcal{T}$ inside the object and $\mathcal{S}$ outside. We can edit the specific object as follows:

\noindent \textbf{Remove.} Removing one object in a hybrid semantic radiance field, actually means that we need to treat the object as transparent in order to expose the background behind the object. For sampling point $t_i \in \mathcal{T}$ of a static object, we set the density $\sigma^s(t_i) = 0$, so that the point will be ignored during volume rendering. As for the dynamic object, we not only set density $\sigma^d(t_i) = 0$ but also set blending weight $b(t_i)=1$. This is because we expect the dynamic object removal operation not to affect the static field.

\noindent \textbf{Filter.} To filter an object from a hybrid semantic radiance field, we set $\sigma(s_i) = 0$ where $s_i \in \mathcal{S}$, which means making all points outside the object invisible.

\noindent \textbf{Composite.} We can composite objects filtered from other scenes(represented by $\mathcal{Z}$) into the current scene by setting $\sigma(s_i) = \sigma(z_i)$, $c(s_i) = c(z_i)$ where $z_i \in \mathcal{Z}$.

\noindent \textbf{Recolor.} For appearance modification, we find editing color $\textbf{c}(s_i)$ in 4D space is the similar to that on 2D images: we exchange RGB channels to change hue parameter, improve the corresponding RGB channel to change RGB saturation parameter, and add all RGB channels to change lightness parameter. This discovery can make it more convenient and controllable for us to recolor. The recoloring results are demonstrated in Fig.~\ref{fig:change_appearance}.

\noindent \textbf{Transform.} We allow users apply various transforming operations to the object, such as translating, scaling, mirroring or duplicating. Users only need to define a transforming function $mapping(\textbf{x}): \textbf{x} \rightarrow \textbf{x}^{\prime}$ to set the spatial or temporal mapping relation of the target object (\textit{e.g.}, $mirror(x,y,z): (-x,-y,z)$, $reverse(x,y,z,t): (x,y,z,-t)$). Then, we set $\sigma(mapping(s_i)) = \sigma(s_i)$, $\textbf{c}(mapping(s_i)) = \textbf{c}(s_i)$ for volume rendering. 
The detailed experimental results can be seen in Fig.~\ref{fig:transformation} .

    

\subsection{Multi-view Reprojection Inpainting}
\label{sec:progressive_inpainting}

\begin{figure}[tbp]
    \centering
    \includegraphics[width=8.5cm, trim=50 170 50 80, clip]{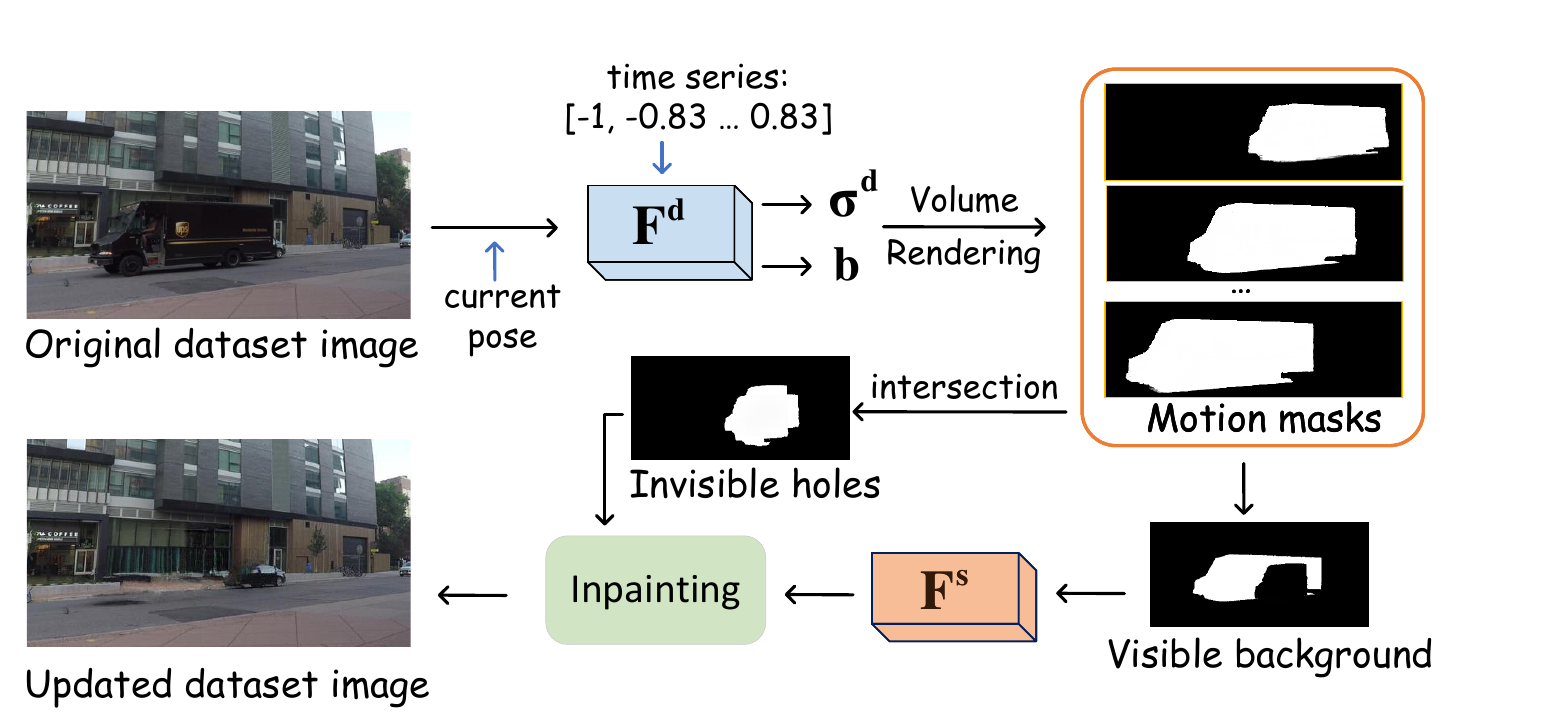}

    \caption{\textbf{Example of fill holes in a image via multi-view reprojection inpainting.}
    }
    \label{fig:reprojection_inpainting}
    \vspace{-4mm}
\end{figure}

Since the limited observations of a scene, the removal operation may cause ``holes'' in the novel views, leading to obvious artifacts.
A feasible solution is to use an inpainting model~\cite{suvorov2022resolution} to fill these holes in edited novel rendered images, and then use these inpainted images as updated training sets to retrain NeRF, similar to the approach employed by SPIn-NeRF\cite{mirzaei2023spin}. However, the inpainting model treats each image as an independent individual, ignoring the multi-view consistency between images, resulting in artifacts and 3D inconsistency. Therefore, we propose multi-view reprojection inpainting method.
As Fig.~\ref{fig:reprojection_inpainting} shows, we divide the training set $\mathcal{J}$ into two parts: $\mathcal{J}_{vis}$ and $\mathcal{J}_{inv}$. The holes in images in the former can be seen in other views. While the holes in images in the latter are invisible across all views. These two parts need to be filled separately.
This belief stems from the notion that, given a specific perspective, if a occluded 3D point is present in some other views, its geometry and appearance information are inherently captured during NeRF's reconstruction, thereby enabling inherent inpainting by NeRF. Conversely, if the occluded points lack observation in other views, then it is necessary to use other models for inpainting.

Therefore, the key objective is to determine the visible and occluded regions within the holes.
To accomplish this, we calculate time-variant motion masks for each individual image in the original dataset. These masks are generated by performing volume rendering on the blending weight $b$ across the entire time series on corresponding camera poses. The overlapping area of these masks represents $\mathcal{J}_{inv}$, while the remaining regions constitute $\mathcal{J}_{vis}$. This process is applied to all original images in the dataset.
For inpainting $\mathcal{J}_{vis}$, we eschew the traditional pixel reprojection method due to its drawbacks such as potential loss of fine details and sensitivity to geometric inconsistencies or occlusions between views.
Instead, we leverage NeRF's inherent multi-view information to accomplish the inpainting of $\mathcal{J}_{vis}$ in all training images. In this way, these holes are filled with rendering results from $F^s$ (Fig.~\ref{fig:reprojection_inpainting}).
Subsequently, lama model~\cite{suvorov2022resolution} is utilized for inpainting on the remaining invisible parts, $\mathcal{J}_{inv}$. 
By pre-filling the background, we narrow down areas that need to be generated, which strengthens the reliability of inpainting results.
Static field $F^s$ is then retrained based on these inpainted images, after which the hybrid semantic radiance field are also retrained. As Fig.~\ref{fig:inpainting_comparison} shows, our proposed inpainting method demonstrates enhanced inpainting results in comparison with only using lama model to fill holes.

\begin{figure*}[tbp]
    \centering
    \includegraphics[width=18cm, trim=80 540 430 80, clip]{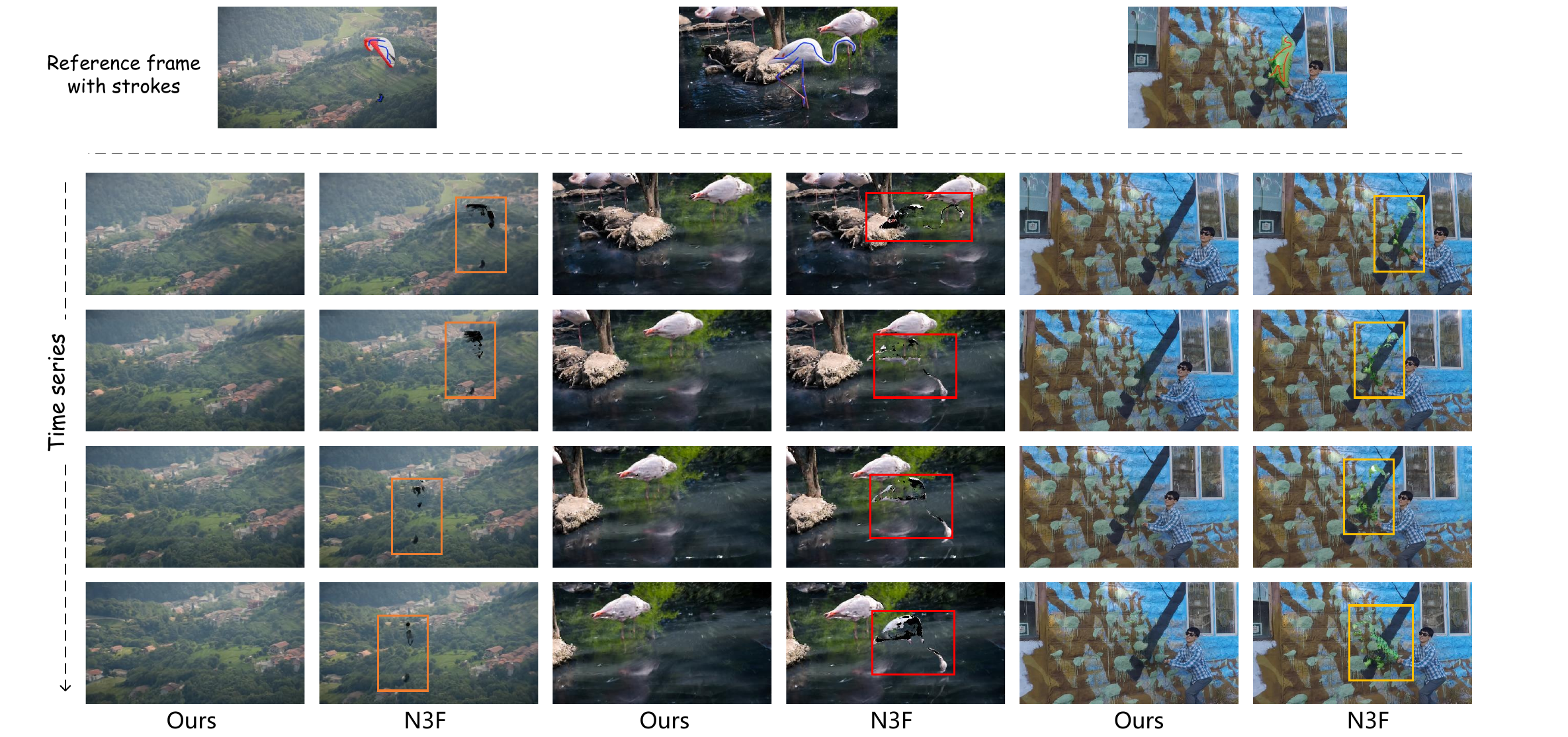}

    \caption{
    \textbf{A Qualitative Comparison of 4D-Editor's Segmentation in 4D space against N3F.}
    Using a constructed feature query based on marked regions in a reference frame, we search for similar regions in the semantic feature fields across the entire time series. Subsequently, the deletion operation affects all novel views accordingly.
    Our results demonstrate that 4D-Editor achieves superior segmentation accuracy compared to N3F, as evidenced by cleaner deletion without artifacts.
    } 
    \label{fig:video_editing}
    \vspace{-4mm}
\end{figure*}

\section{Experiments}
\label{sec:experiments}

\subsection{Experimental setup}
\label{sec:experiment_setup}
\textbf{Datasets.} We experiment on three datasets: Dynamic View Synthesis\cite{yoon2020novel}, DAVIS\cite{Perazzi_CVPR_2016} and NeuPhysics~\cite{qiao2022NeuPhysics }.

\textbf{Implements Details.} We implement 4D-Editor with Pytorch, using Adam optimizer to update learnable parameters on one NVIDIA A6000 GPU. 
We train original hybrid NeRF and semantic parts separately, where training DynNeRF or RobustNeRF can take 6-8 hours, while time for training additional distillation of hybrid semantic features is only 10-15 minutes.
Editing one frame can take 1-2 seconds based on RobustNeRF and 8 seconds based on DynNeRF.


\subsection{Interactive Object Removal using strokes}
\label{sec:experiment_segmentation}
\begin{figure}[tbp]
    \centering
    \includegraphics[width=8cm, trim=40 150 240 180, clip]{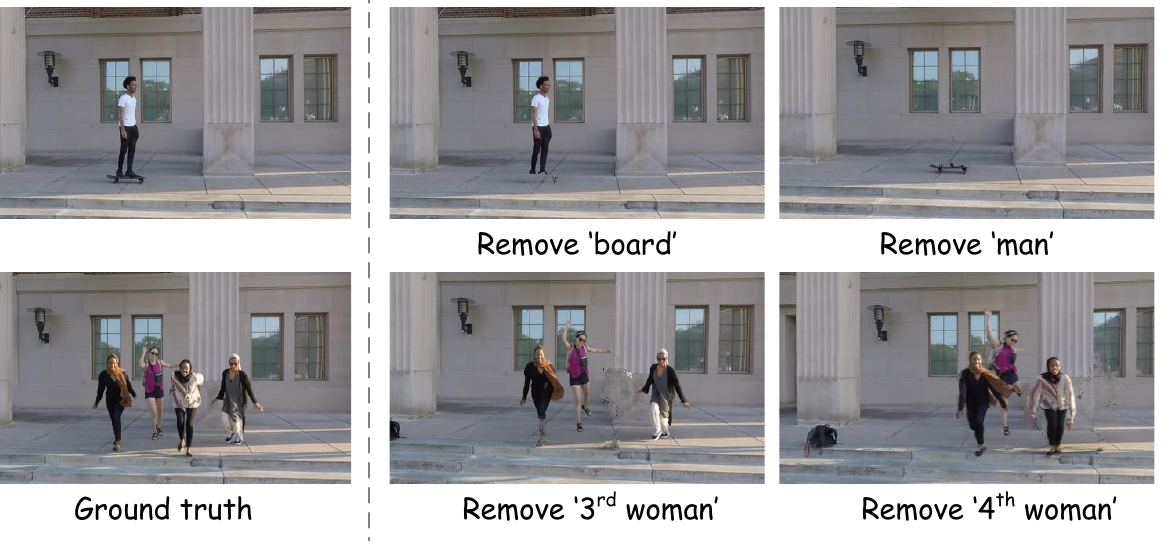}
    \caption{
    \textbf{Object-Level Editing.}
    We individually remove different objects within the same scene, and only the user-defined target area is affected, with rest regions remaining unaffected.
    } 
    \label{fig:remove_different_parts}
    \vspace{-4mm}
\end{figure}

Our method enables users to perform interactive object-level editing with strokes.
As Fig.~\ref{fig:video_editing} shows, the user simply annotates the target object on a reference frame of the original videos(Row 1). After constructing the editing command (\textit{e.g.}, removal), this operation can be propagated to the whole scenes. Compared with N3F~\cite{tschernezki2022neural}, our method can achieve clean and continuous removal effects, whereas N3F leaves obvious artifacts. Since ISRF\cite{goel2023interactive} cannot edit dynamic scenes and NeuPhysics\cite{qiao2022NeuPhysics }'s code on editing module is not available, we make no comparison with them.


Our proposed 4D-Editor can remove arbitrary objects in dynamic NeRF by applying different queries to match with semantic fields, while maintaining spatial-temporal consistency in rest regions as shown in Fig.~\ref{fig:remove_different_parts}. Moreover, 4D-Editor also enables users to edit multiple objects, such as deleting the women, recoloring, and shifting the balloon simultaneously, as demonstrated in Fig.~\ref{fig:head}, which is achieved by constructing multiple queries and editing commands at one time.




\subsection{Refined and Diverse Editing}
\label{sec:experiment_diverse_editing}
\begin{figure}[tbp]
    \vspace{-3mm}
    \centering
    \subfloat[saturation $\downarrow$]{
        \includegraphics[width=0.09\textwidth, trim=280 80 20 30, clip]{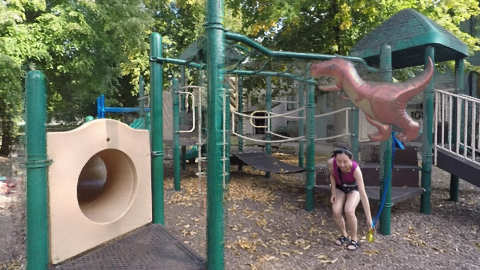} 
    }
    \subfloat[saturation $\uparrow$]{
        \includegraphics[width=0.09\textwidth, trim=280 80 20 30, clip]{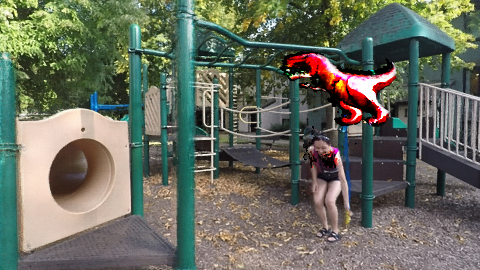} 
    }
    \subfloat[lightness $\downarrow$]{
        \includegraphics[width=0.09\textwidth, trim=250 80 50 30, clip]{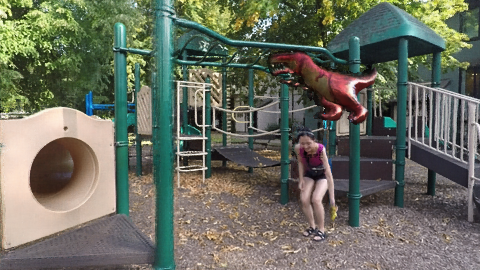} 
    }
    \subfloat[lightness $\uparrow$]{
        \includegraphics[width=0.09\textwidth, trim=250 80 50 30, clip]{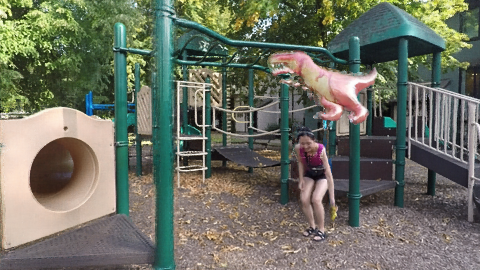} 
    }
    \vspace{1mm}
    \caption{\textbf{Recolor the balloon.}} 
    \label{fig:change_appearance}
    \vspace{-3mm}
\end{figure}

4D-Editor facilitates accurate segmentation to achieve controllable modifications in appearance color, akin to the HSL (Hue, Saturation, and Lightness) editing in PhotoShop\cite{photoshop}. As illustrated in Fig.~\ref{fig:change_appearance}, we begin by selecting the balloon, which is initially red in color, and modify its hue, saturation, and lightness individually. This is accomplished by increasing or decreasing values in all color channels.

\begin{figure}[tbp]
    \vspace{-3mm}
    \centering
    \subfloat[composition]{
        \includegraphics[width=0.14\textwidth]{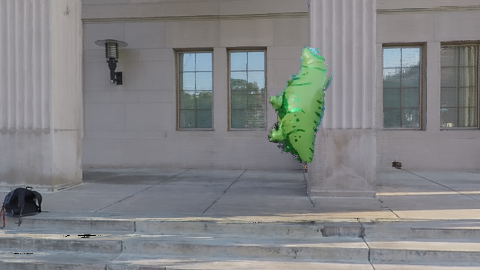}
        \label{fig:composition}
    }
    \subfloat[mirror]{
        \includegraphics[width=0.14\textwidth]{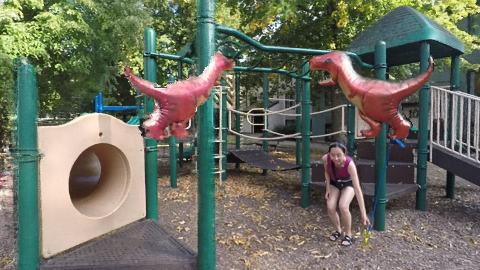}
        \label{fig:transform_mirror}
    }
    \subfloat[scale]{
        \includegraphics[width=0.14\textwidth]{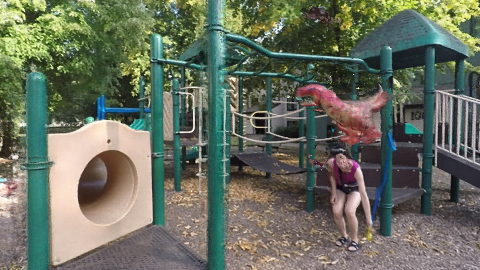}
        \label{fig:transform_scale}
    }
    \vspace{1mm}
    \caption{
    \textbf{Composition \& Transformation.}
    4D-Editor supports composition across different scenes and flexible transformations(More results in \textcolor{blue}{supplementary material Sec. F}).
    } 
    \label{fig:transformation}
    \vspace{-3mm}
\end{figure}

Additionally, 4D-Editor provides users with the flexibility to define their own transformation functions to apply affine transformations on individual selected objects.
Fig.~\ref{fig:transformation} illustrates the effects of object-level composition across multiple scenes and a variety of transformations.
The green balloon is actually filtered from another scene and inserted into the playground in Fig.~\ref{fig:composition}.
As for different transformation operations, we can still keep the correct spatial information: The mirrored balloon is partially traversed in Fig.~\ref{fig:transform_mirror} while previously obscured background become visible after scaling in Fig.~\ref{fig:transform_scale}.

\subsection{Ablation Studies}
\label{sec:ablation}
\noindent \textbf{Recursive Selection Refinement.} 
For accurate object segmentation in hybrid semantic radiance field, it is essential to utilize a 2D-4D feature matching technique that matches marked features with the distilled semantic features in volumetric space.
As depicted in Fig.~\ref{fig:ablation_refinement}, we evaluate three different methods: (1) Average feature, (2) K-Means clustered features \cite{goel2023interactive}, and (3) Recursive selection refinement (our method).
Fig.~\ref{fig:ablation_average_feature} illustrates ineffective feature matching resulting from the limited capability of average features to extract meaningful features.
K-Means clustered features only remove part of object as shown in Fig.~\ref{fig:ablation_depth_1}, leading to artifacts(\textit{e.g.}, remaining legs).
However, our proposed recursive selection refinement improves the precise of feature matching and achieves near perfect removal, as demonstrated in Fig.~\ref{fig:ablation_depth_3}-\ref{fig:ablation_depth_30}.



Table~\ref{table:points_in_each_iteartion} presents the number of newly added valid points and the total number of possible points in each recursion. We can see that when the recursive number reaches $20$, almost all points belonging to the target is selected and ultimately converged to $3.24M$.
We also studied the effect of different recursive number on IoU and Acc in Table~\ref{table:iou_acc}. We can find that as the number of recursive number $K$ increases, the object segment accuracy also improves, ultimately converging to around $93\%$.
(See more analysis in \textcolor{blue}{supplementary material Sec. E}).

\begin{table}[tbp]
\footnotesize
\centering
\setlength{\tabcolsep}{1mm}{
\renewcommand{\arraystretch}{1.2}
\begin{tabular}{| c | c c c| c | c c c|}
\hline
\textbf{$K$} & \textbf{$new\ \mathcal{V}$} &  \textbf{$\mathcal{P}$} & \textbf{$all\ \mathcal{V}$} & 
\textbf{$K$} & \textbf{$new\ \mathcal{V}$} &  \textbf{$\mathcal{P}$} & \textbf{$all\ \mathcal{V}$} 
\\
\hline
1     & 2.84\;M & 828.13\;k & 2.84\;M &    10 & 3.68\;k & 35.58\;k & 3.22\;M\\ 
2     & 225.93\;k & 373.77\;k & 3.06\;M &    20 & 426 & 6.55\;k & 3.23\;M\\ 
3     & 63.97\;k & 239.31\;k & 3.13\;M &    30 & 114 & 1.64\;k & 3.24\;M\\ 
5     & 20.69\;k & 120.73\;k & 3.18\;M &    50 & 8 & 169 & 3.24\;M \\ 
8     & 6.73\;k & 54.89\;k & 3.21\;M &    60 & 1 & 56 & 3.24\;M \\ 
\hline
\end{tabular}}
\vspace{2mm}
\caption{
\textbf{Points in each recursion iteration.}
When the recursive number reaches $20$, the number of new valid points decreases significantly ($\alpha=0.6, \beta=0.1$), which converges to 3.24M despite further recursion.
}
\label{table:points_in_each_iteartion}
\vspace{-3mm}
\end{table}

\begin{table}[tbp]
\footnotesize
\centering
\setlength{\tabcolsep}{1.5mm}{
\renewcommand{\arraystretch}{1.2}
\begin{tabular}{| c | c c| c | c c|}
\hline
\textbf{$K$} & \textbf{$IoU$} &  \textbf{$Acc$} &
\textbf{$K$} & \textbf{$IoU$} &  \textbf{$Acc$} 
\\
\hline
1     & 72.79\;\% & 79.81\;\%  &    30 & 78.63\;\% & 93.45\;\%\\ 
5     & 76.53\;\% & 85.39\;\%  &    40 & 80.11\;\% & 93.24\;\% \\ 
20     & 80.85\;\% & 92.70\;\% &    50 & 79.21\;\% & 93.88\;\% \\ 
\hline
\end{tabular}}
\vspace{1mm}
\caption{
\textbf{Effects of maximum recursion number $K$.}
}
\label{table:iou_acc}
\vspace{-3mm}
\end{table}

\begin{figure}[tbp]
\centering
    \includegraphics[width=8cm, trim=100 180 260 280, clip]{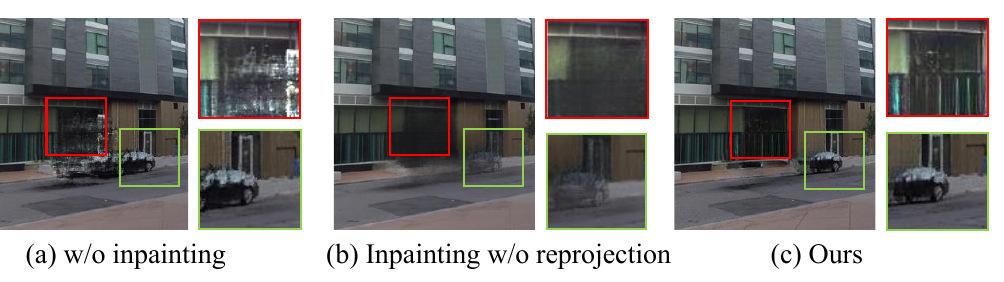}
    \caption{\textbf{Comparison of different inpainting methods.}
    To fill holes in \textbf{(a)}, we pre-fill backgrounds through reprojection and narrow the inpainting region for reliable results(\textit{e.g.}, the black car remains in \textbf{(c)} while disappears in \textbf{(b)}).
    }
    \label{fig:inpainting_comparison}
    \vspace{-4mm}
\end{figure}

\begin{figure}[tbp]
\centering
    \includegraphics[width=8cm, trim=80 450 80 450, clip]{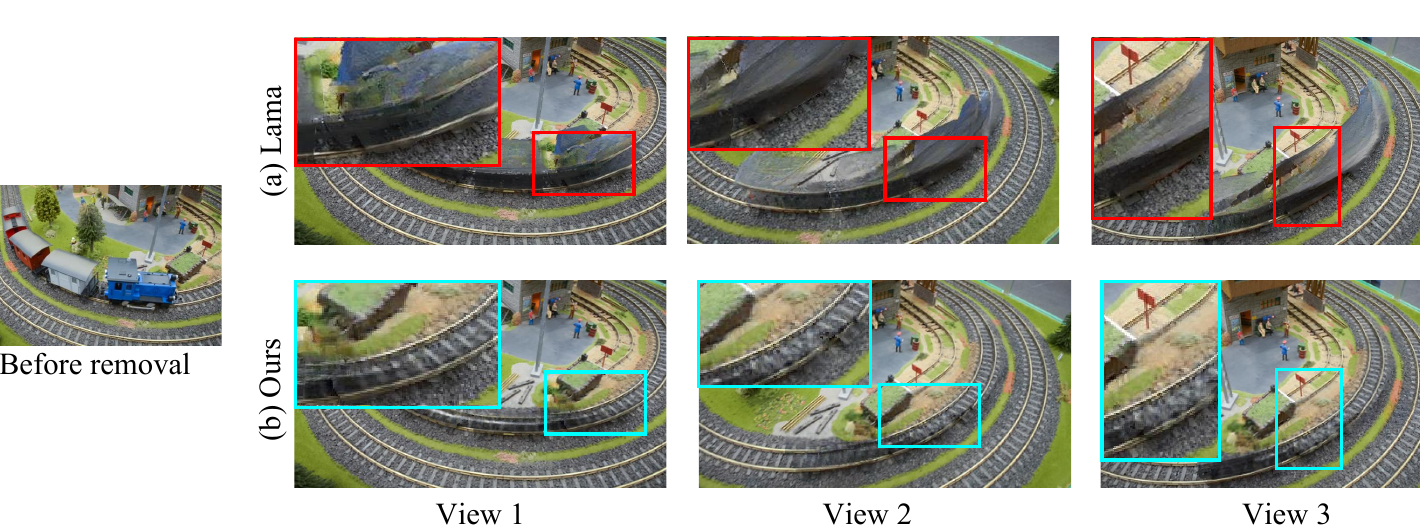}
    \caption{\textbf{Visulaizations of multi-view consistency.}
    The top raw demonstrates that direct inpainting causes view-inconsistent with obvious artifacts, while our method (the bottom raw) exhibits better multi-view consistency.
    }
    \label{fig:inpainting_multiview}
    \vspace{-4mm}
\end{figure}

\noindent \textbf{Multi-view Reprojection Inpainting.}
By employing multi-view reprojection, we restore backgrounds using information captured from multiple perspectives (refer to Fig.~\ref{fig:inpainting_comparison}).
In contrast to the direct inpainting method, which results in the loss of significant spatial information (\textit{e.g.}, the disappearance of the black car in Fig.~\ref{fig:inpainting_comparison}\textcolor{red}{b}) and produces independent multi-view outcomes causing blurry artifacts after re-training $F^s$, our method preserves the original spatial information in the scene.
Furthermore, after pre-filling in the `visible' parts of the backgrounds by reprojection, we narrow down the inpainting areas to alleviate the failure caused by inpainting models, obtaining more reliable and detailed outcomes (\textit{e.g.}, walls with clear textures in Fig.~\ref{fig:inpainting_comparison}\textcolor{red}{c}).
Fig.~\ref{fig:inpainting_multiview}) demonstrates that our proposed multi-view reprojection inpainting method can achieve better multi-view consistency compared with only using inpainting models to fill holes. Moreover, the inpainting performance can be further improved by employing more powerful inpainting models.

\section{Conclusion}
\label{sec:conclusion}
We propose a novel interactive editing framework for dynamic scenes that enables object-level editing operations through user-provided strokes on a single reference frame, and delivers spatial-temporal consistency across the entire time series.
We present several excellent results from multiple challenging scenes.
However, our method has limitations in removing shadows of moving objects. Additionally, while we can handle invisible background completion, in some times, the scene inpainting may still have spatial-temporal inconsistencies. We will investigate ways to address the problem in future works.

{\small
\bibliographystyle{ieee_fullname}
\bibliography{egbib}
}

\appendix


\twocolumn[{
\maketitle
\begin{figure}[H]
\hsize=\textwidth 
\centering
\vspace{-45mm}
\includegraphics[width=12cm, trim=100 120 100 20, clip]{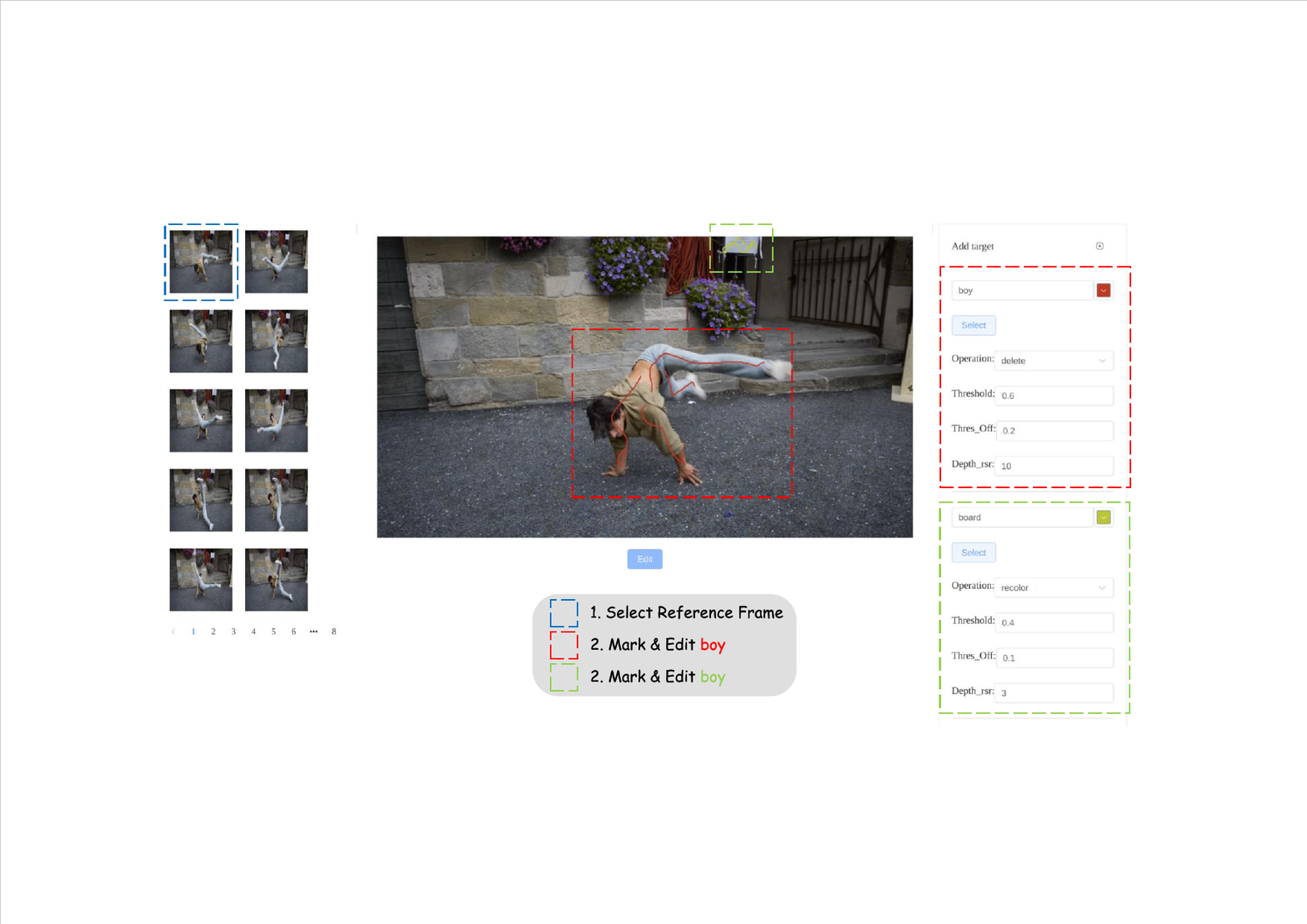}
\caption{
\textbf{GUI.}
The 4D-Editor provides users with a graphical user interface (GUI) to interactively edit dynamic scenes. Users first select a reference frame available in the left column and mark multiple objects using strokes in distinct colors. Once required parameters are specified to generate feature queries and editing commands (e.g., \textbf{delete} boy [\textcolor{red}{$\sim$}] and \textbf{recolor} board [\textcolor{green}{$\sim$}], the editing effects propagate throughout the entire time series, ensuring that all novel views are appropriately modified. The process can be found on \href{https://patrickddj.github.io/4D-Editor}{project page}.
}
\label{fig:gui}
\end{figure}
}]

\section{Implementation Details}
We use Principal Component Analysis (PCA) method to extract 64 most important semantic features from DINO~\cite{caron2021emerging} ViT-b8 model. We use 64 sampled points on each ray in volume rendering. 

During the editing process, we set different threshold $\alpha$ and number of K-Means clusters $N_k$ according to different sizes of target objects: $\alpha = 0.4\pm 0.1$, $N_k = 3 \pm 2$ for small objects and $\alpha = 0.7\pm 0.1$, $N_k = 20 \pm 15$ for small ones.
The value of $\beta$ should be approximately 1/3 of $\alpha$.
Actually, in our experiments, all these values can be set loosely within Recursive Selection Refinement.

\section{Interactive GUI}
We design a user-friendly graphical user interface (GUI) that facilitates interactive editing of dynamic scenes. The editing procedure involves three steps: 1) Selecting a reference frame, 2) Marking target objects with strokes in distinct colors, and 3) Configuring editing parameters, including editing operations, threshold $\alpha$, exploration range $\beta$ and recursion depth $K$. These steps are depicted in Fig.~\ref{fig:gui}.

\section{Model Structure}

\subsection{Structures of Hybrid Radiance Fields}
\begin{figure}[hbp]
    \centering
    \subfloat[Robust NeRF~\cite{liu2023robust}]{\includegraphics[width=8cm, trim=0 0 0 0, clip]{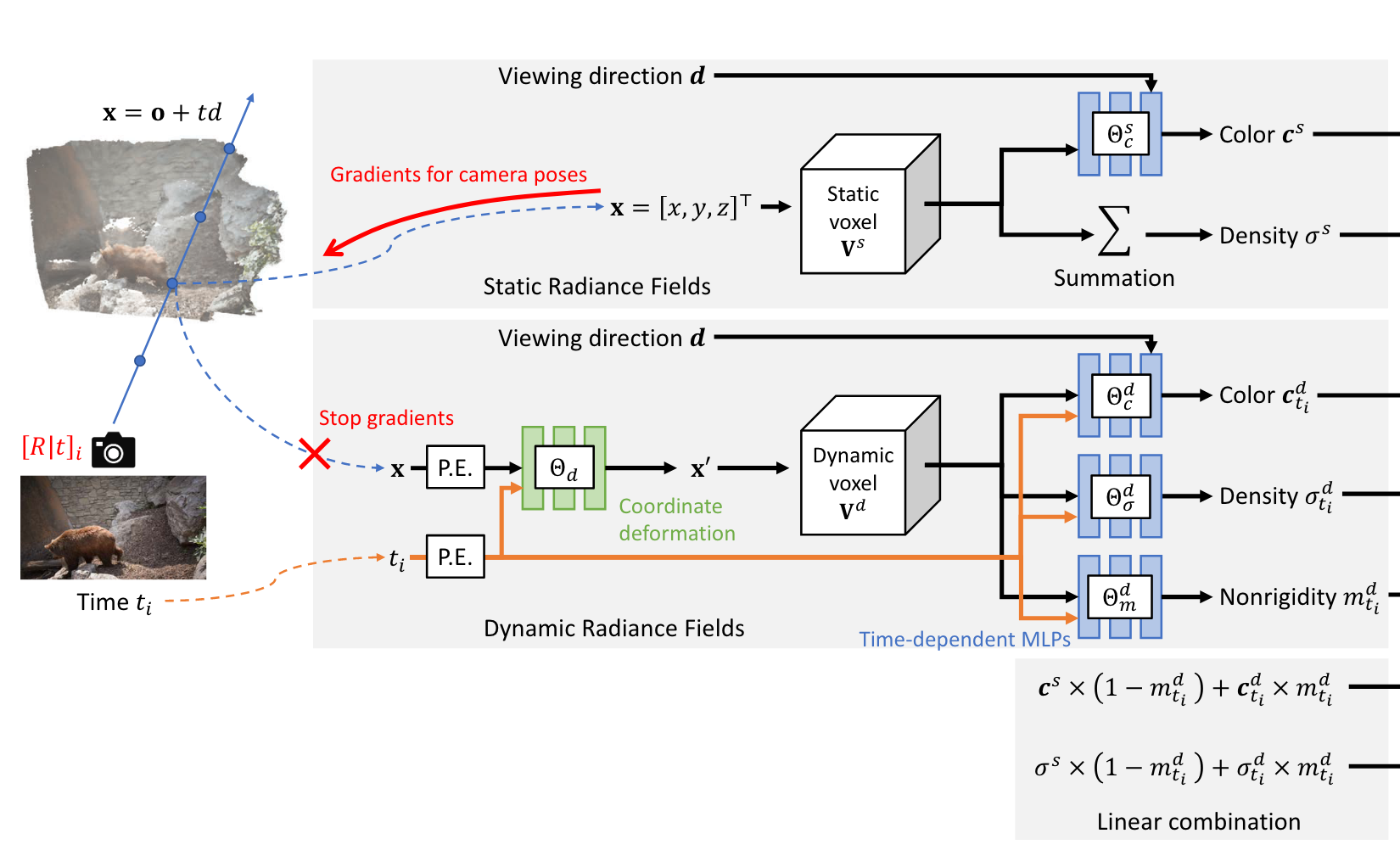}}
    \\
    \subfloat[Dynamic NeRF~\cite{yoon2020novel}]{\includegraphics[width=8cm, trim=0 0 0 0, clip]{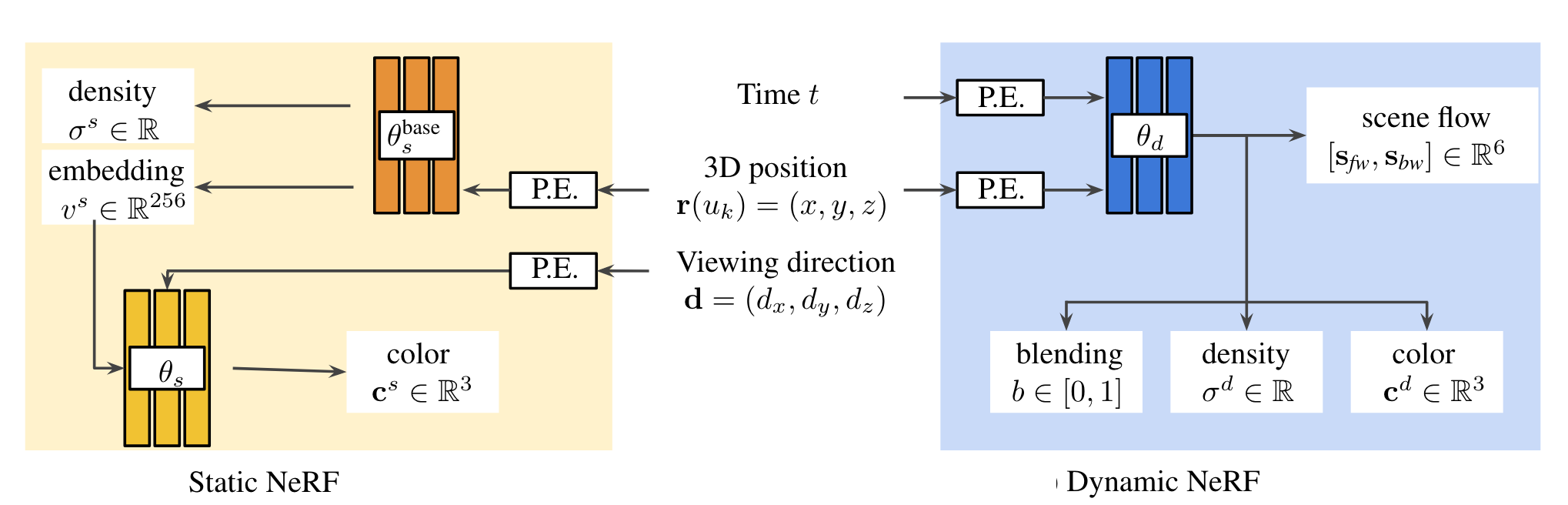}}
    \vspace{1mm}
    \caption{
    \textbf{Hybrid Radiance Field Reconstruction of Dynamic Scenes.} The structures of these two methods are directly adopted from their respective theses.
    } \label{fig:model_hybrid}
\end{figure}

We directly adopt RobustNeRF~\cite{liu2023robust} and DynamicNeRF~\cite{yoon2020novel} to represent a dynmaic scene, their methods are displayed in Fig.~\ref{fig:model_hybrid}.

\subsection{Structures of Hybrid Semantic Fields}
\begin{figure}[ht]
    \centering
    \subfloat[Static Feature Field $G^s$]{\includegraphics[width=6cm, trim=180 830 10 100, clip]{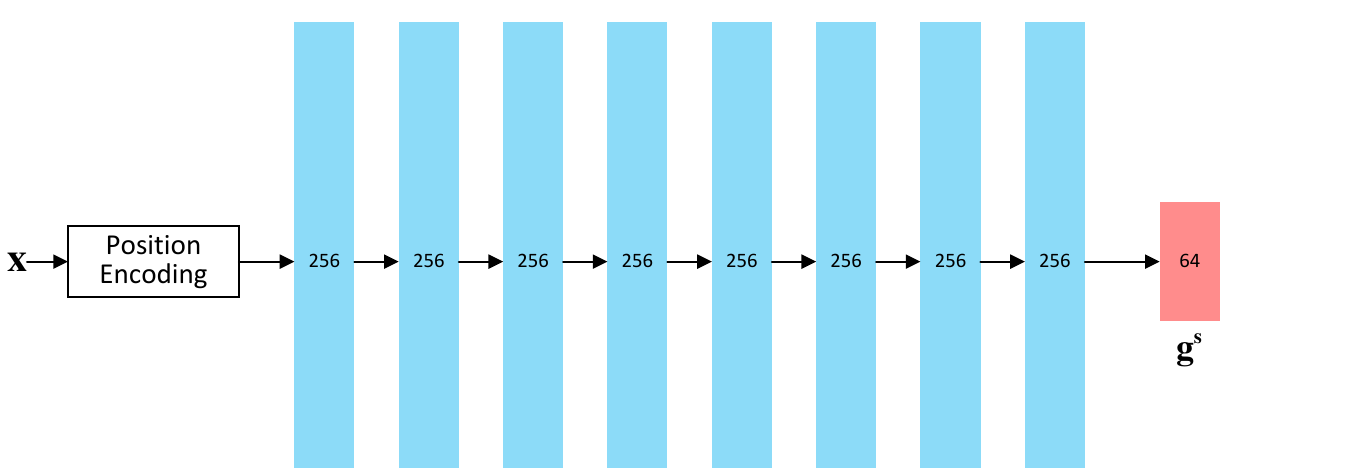}}
    \\
    \subfloat[Dynamic Feature Field $G^d$]{\includegraphics[width=6cm, trim=180 530 10 400, clip]{images/model_structure.pdf}}
    \vspace{1mm}
    \caption{
    \textbf{Structures of Semantic Fields.}
    }
    \label{fig:model_structure}
\end{figure}

We design semantic fields $G^s$ and $G^d$, to represent semantic information of the static and dynamic components in the scene, respectively. Both semantic fields are modeled using an 8-layer multi-layer perceptron (MLP). As illustrated in Fig.~\ref{fig:model_structure}, $G^s$ takes the position $\textbf{x}$ as input after position encoding, while $G^d$ incorporates both the position $\textbf{x}$ and the time $t$.

\subsection{Volume Rendering on Semantic Features}
\begin{figure}
    \centering
    \includegraphics[width=8cm, trim=10 230 300 10, clip]{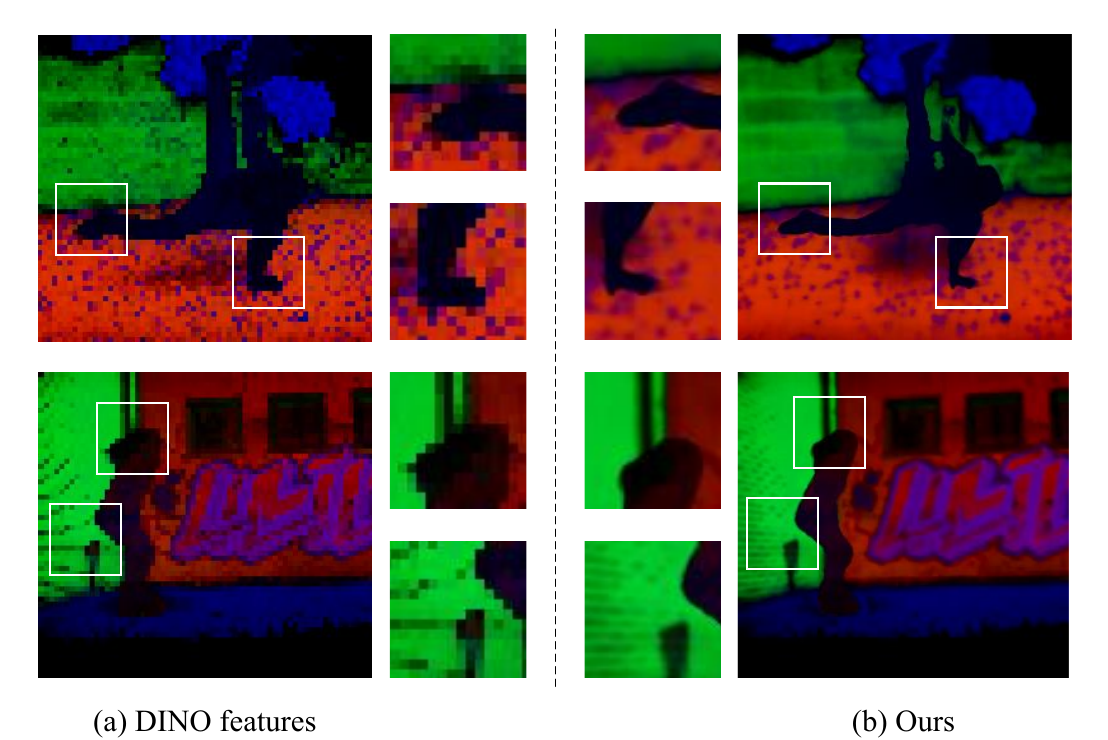}
    \caption{
    \textbf{Visualization of Semantic Features.}
    After semantic features distillation, we obtain enhanced semantic features in 4D space (For additional comparisons, refer to the \href{https://patrickddj.github.io/4D-Editor}{project page}).
    }
    \label{fig:feature_rendering}
    \vspace{-3mm}
\end{figure}

The spatial semantic features themselves are represented by a 64-dimensional vector. We apply volume rendering to these features and select the first 3 dimensions to generate RGB visualizations. 
As shown in Fig.~\ref{fig:feature_rendering}, upon reconstructing the spatial semantic information using semantic fields, our approach demonstrates superior preservation of both multi-view and spatio-temporal consistency of semantic information in the 4D space, compared to the relatively coarse-grained feature map generated by the DINO model~\cite{caron2021emerging}. 
Notably, on object edges, the transition of semantic information appears remarkably smooth.

\section{Visualization of Recursive Selection Refinement}
\begin{figure}[ht]
    \centering
    \includegraphics[width=8.5cm, trim=300 300 150 100, clip]{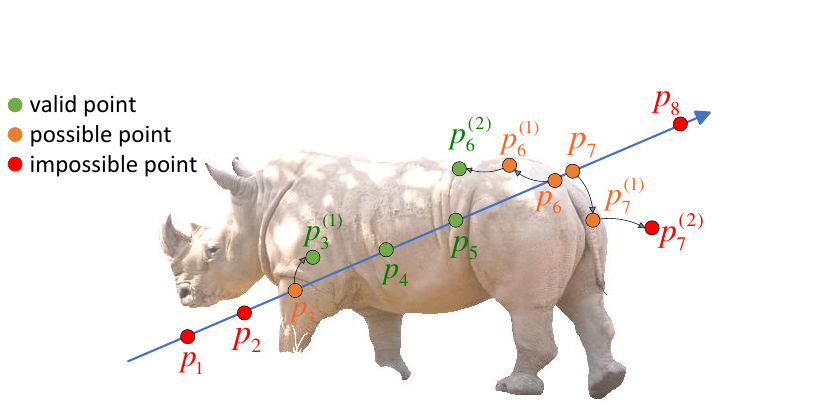}
    \caption{
    \textbf{Process of Recursive Selection Refinement.}
    Although DINO model generates coarse semantic feature maps which are used as the training set, our reconstruction and volume rendering of the 4D spatial semantic information results in finer semantic feature maps.
    }
    \label{fig:main_recursive_selection_refinement}
    \vspace{-2mm}
\end{figure}

Figure~\ref{fig:main_recursive_selection_refinement} shows the flow of semantic feature matching and target selection(semantic segmentation).
Given sampled points $p_1 \dots p_8$, we categorize them based on feature distances: valid points \textcolor[RGB]{69 139 0}{$p_4\;p_5$}, impossible points \textcolor{red}{$p_1\;p_2\;p_8$}, possible points \textcolor{orange}{$p_3\;p_6\;p_7$}.
Then we introduce a random offset on possible points and recalculate feature distances, repeating until these points are judged as valid or impossible.
Here, \textcolor{orange}{$p_3\;p_6$} are considered as valid points after 1 or 2 iterations respectively, while \textcolor{orange}{$p_7$} excluded after 2 iterations.
Ultimately, the selected points are \textcolor{orange}{$p_3$}, \textcolor[RGB]{69 139 0}{$p_4$}, \textcolor[RGB]{69 139 0}{$p_5$}, \textcolor{orange}{$p_6$}.
We will repeat this process for all sampling points on each ray during the rendering process.
As for the same batch of sampling points, we can apply different random offset step $s$ at one time (\textit{e.g.}, $s = (1/50, 1/100, 1/150)$) to control different ranges of selection on target.

\begin{figure*}[tbp]
    \centering
    \includegraphics[width=16cm, trim=10 120 10 200, clip]{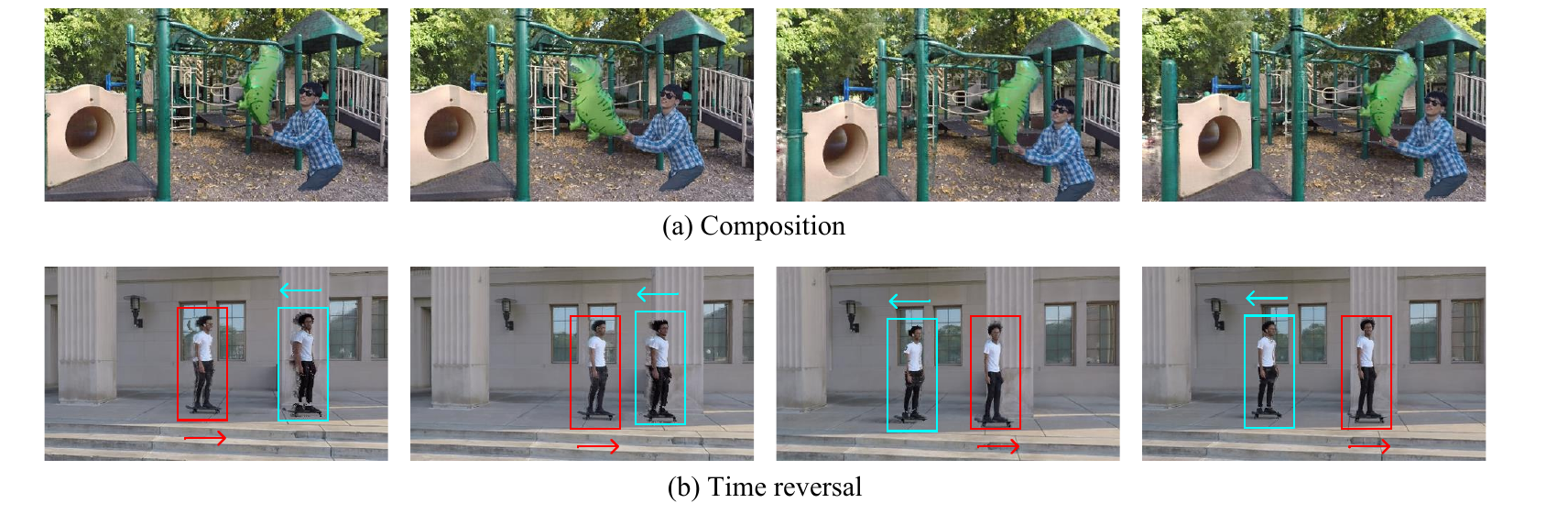}
    \caption{
    \textbf{Composition \& Time Reversal.}
    We maintain spatial-temporal consistency across the entire time series when combining different scenes or applying time-variant transformations to individual objects within the same scene. See more on \href{https://patrickddj.github.io/4D-Editor}{project page}.
    }
    \label{fig:composition+reversal}
\end{figure*}

\section{Analysis on Object Segmentation}
In this section, we begin by conducting a qualitative and quantitative comparison between N3F~\cite{tschernezki2022neural} and our method in terms of object segmentation performance. Subsequently, we evaluate the impact of the hyperparameter $\beta$ (exploration range) on Recursive Selection Refinement.

\subsection{Evaluation on Segmentation Accuracy}
\begin{figure}[tbp]
    \centering
    \includegraphics[width=9cm, trim=80 320 320 10, clip]{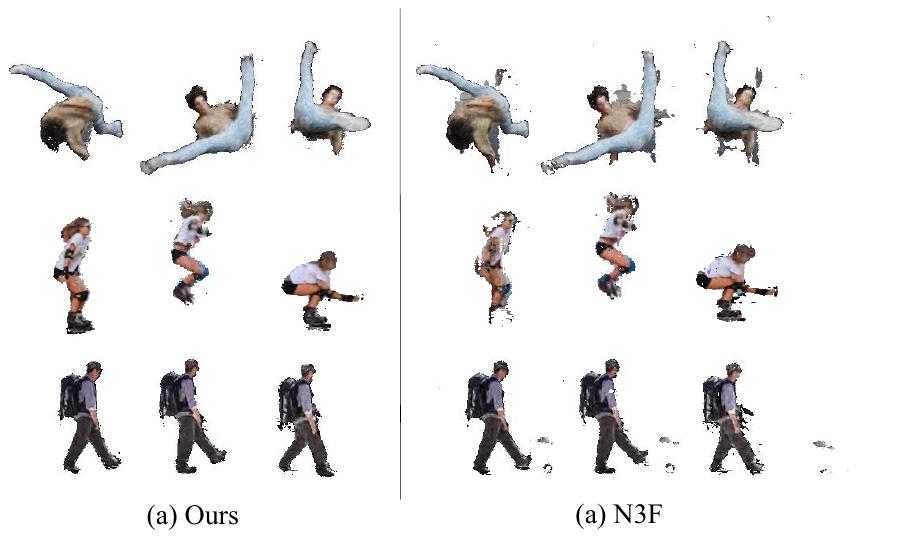}
    \caption{\textbf{Results of Object Segmentation.} 
    Compared with N3F, we achieve finer object segmentation, which contains fewer artifacts especially for edge areas.
    }
    \label{fig:segmentation_davis}
    \vspace{-1mm}
\end{figure}

\begin{figure}[tbp]
    \centering
    \subfloat[balloon]{\includegraphics[width=2.55cm, trim=260 70 60 30, clip]{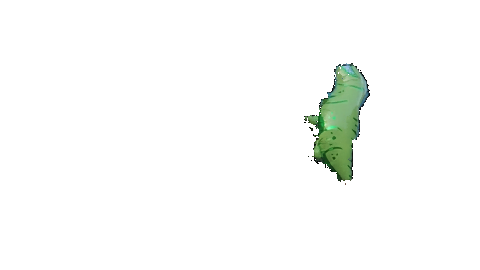}}
    \subfloat[woman]{\includegraphics[width=2.55cm, trim=290 10 50 120, clip]{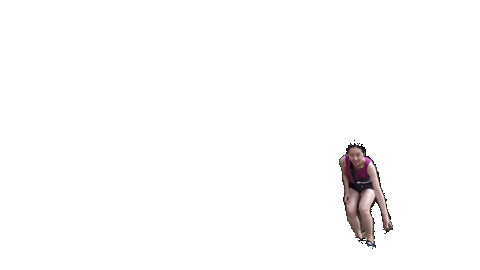}}
    \subfloat[balloon]{\includegraphics[width=2.55cm, trim=250 100 70 10, clip]{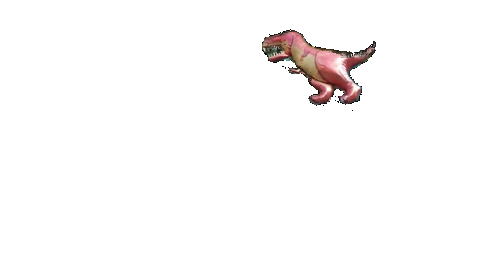}}
    \vspace{1mm}
    \caption{
    \textbf{Object Segmentation in 4D Space.}
    } \label{fig:filter_balloon}
\end{figure}

We fine-tune the threshold of N3F for object segmentation tasks and dispaly its best results in the right column of Fig.~\ref{fig:segmentation_davis}. Since our method utilizes Recursive Selection Refinement to achieve more precise selection of target objects, surpassing N3F in both qualitative and quantitative assessments (indicated by higher Acc and IoU values in Table.~\ref{table:segmentation_davis}).
In our experiments, we observe that our method performs especially well in scenes containing significant object movement (\textit{e.g.}, a girl starting from the left side and jumping to the right side in the \textit{Rollerblade} dataset). Conversely, N3F produces numerous broken artifacts in such scenes.
Fig.~\ref{fig:filter_balloon} presents additional results of object segmentation in 4D space.

\subsection{Analysis on exploration range $\beta$}
The selection of the exploration range parameter $\beta$ can be loose and flexible. Fig.~\ref{table:exploration_range} demonstrates that the segmentation quality remains consistent despite using different values of $\beta$.
This finding supports the notion that it is the recursion depth parameter $K$ that significantly enhances segmentation accuracy, while $\beta$ merely serves as an auxiliary factor.

\begin{table}[tbp]
\footnotesize
\centering
\setlength{\tabcolsep}{4.5mm}{
\renewcommand{\arraystretch}{1.4}
\begin{tabular}{| c | c | c | c |}
\hline
\textbf{Scene} & \textbf{Metric} &  \textbf{N3F} & \textbf{Ours} \\ \hline
\multirow{2}{*}{\textit{Breakdance Flare}}  & \textbf{Acc $\uparrow$} & 89.96\% & \textbf{94.12\%} \\ \cline{2-4}
                                            & \textbf{IoU $\uparrow$} & \textbf{84.97\%} & 83.09\% \\ \hline
\multirow{2}{*}{\textit{Rollerblade}}       & \textbf{Acc $\uparrow$} & 83.38\% & \textbf{93.10\%} \\ \cline{2-4}
                                            & \textbf{IoU $\uparrow$} & 71.55\% & \textbf{81.56\%} \\ \hline
\multirow{2}{*}{\textit{Hike}}              & \textbf{Acc $\uparrow$} & 94.80\% & \textbf{95.16\%} \\ \cline{2-4}
                                            & \textbf{IoU $\uparrow$} & 85.73\% & \textbf{89.88\%} \\ \hline
\end{tabular}}
\vspace{2mm}
\caption{
\textbf{Quantitative Analysis on Object Segmentation.} We choose several scenes from DAVIS dataset and use its true motion masks as ground truth.
}
\label{table:segmentation_davis}
\vspace{-3mm}
\end{table}

\begin{table}[tbp]
\footnotesize
\centering
\setlength{\tabcolsep}{1.1mm}{
\renewcommand{\arraystretch}{1.2}
\begin{tabular}{| c | c c c c c c c|}
\hline
\textbf{$\beta$} & 0.05 &  0.08 & 0.1 & 0.12 & 0.15 & 0.18 & 0.2 
\\
\hline
\textbf{Acc $\uparrow$} & 93.10\% & 93.42\% & 93.81\% & 93.66\% & 93.94\% & 94.02\% & 93.94\% 
\\ 
\textbf{IoU $\uparrow$} & 81.56\% & 81.26\% & 80.90\% & 80.77\% & 80.21\% & 79.65\% & 79.26\%
\\ 
\hline
\end{tabular}}
\vspace{2mm}
\caption{
\textbf{Effects of exploration range $\beta$.} We set the parameters as follows: $K=3$, $N_k=8$, and $\alpha=0.6$. Additionally, we experiment with different values of $\beta$ to demonstrate that it has no impact on the segmentation quality. Consequently, $\beta$ can be loosely set during editing operations.
}
\label{table:exploration_range}
\vspace{-2mm}
\end{table}

\section{Scene Editing}
This section focuses on two complex editing operations: transformation and composition(Fig.~\ref{fig:composition+reversal}).

\subsection{Transformation Details}
Table~\ref{table:trans_func} shows three types of geometric transformation operations (\textit{Shift}, \textit{Scale}, and \textit{Mirror}) and a time-variant operation (\textit{Reverse}). The time-variant transformation allows editing in the temporal dimension. For example, we can reverse the trajectory of moving objects in a time series while maintaining the others (\textit{e.g.}, Fig.~\ref{fig:composition+reversal}\textcolor{red}{b} shows two boys: the boy in the red box follows the real trajectory to the right side, but the boy in the blue box moves towards the opposite direction and reverses to the left side).

In dynamic scenes, we lack explicit knowledge regarding the distribution of different objects in both the current space and the space after transformation. We can only judge the selected target region by calculating the feature distance. Therefore, we perform two rounds of calculations for all sampled points. In the first round, we use initial positions of all sampled points, dividing them into $\mathcal{T}$ (inside the object) and $\mathcal{S}$ (outside) according to Section \textcolor{red}{3.3}. In the second round, we use the positions after transformation, dividing edited space into $\mathcal{T}^{\prime}$ and $\mathcal{S}^{\prime}$. Additionally, for each sampled point $u_i$, we obtain its original attributes $c^o$ and $\sigma^o$, as well as $c^t$ and $\sigma^t$ after transformation.

To preserve the original object, it is kept intact. Alternatively, if removal is desired, we set $\sigma = 0$. In cases where there are overlapping areas between the transformed object and the original object, we consistently apply $c^t$ and $\sigma^t$ to assign the properties to these points. The settings for the final density and color can be found in Table~\ref{table:after_trans} provided below.

\begin{table}[tbp]
\footnotesize
\centering
\setlength{\tabcolsep}{1.5mm}{
\renewcommand{\arraystretch}{1.5}
\begin{tabular}{| c | c |}
\hline
\textbf{Transformation} & \textbf{Mapping Function}
\\
\hline
\textit{Shift}     &  $Shift(x, y, z): (x, y, z) + \delta$ \\ 
\textit{Scale}     & $Scale(x, y, z): (x, y, z) \times scale$ \\ 
\textit{Mirror}     & $Mirror(x, y z): (-x, -y, -z)$ \\ 
\textit{Reverse}     & $Reverse(x, y, z, t): (x, y, z, -t)$\\
\hline
\end{tabular}}
\vspace{2mm}
\caption{
\textbf{Transformation Functions.}
}
\label{table:trans_func}
\end{table}

\begin{table}[tbp]
\footnotesize
\centering
\setlength{\tabcolsep}{3.3mm}{
\renewcommand{\arraystretch}{1.5}
\begin{tabular}{| c | c |}
\hline
$u_i \in \mathcal{T}^{\prime}$     
&  $(\sigma, c) \leftarrow (\sigma^t c^t)$ \\ 
\hline
$u_i \in \mathcal{T}, u_i \notin \mathcal{T}^{\prime}$
& $(\sigma, c) \leftarrow (\sigma^{o} c^{o})$ if reserved \\
& $\sigma \leftarrow 0$ if not reserved \\
\hline
$u_i \notin \mathcal{T}, u_i \notin \mathcal{T}^{\prime}$ & $(\sigma, c) \leftarrow (\sigma^o c^o)$
\\
\hline
\end{tabular}}
\vspace{2mm}
\caption{
\textbf{Final $\sigma$, $c$ after Transformation.}
}
\label{table:after_trans}
\vspace{-3mm}
\end{table}

\subsection{Composition Details}
Assuming the presence of $N$ objects derived from distinct dynamic scenes, which have been filtered from the original NeRF and are represented using masks in 4D space.
Transformation can also be applied here to prevent object overlap during composition.
The composition is performed at each sample point for all segmented parts collectively:
\begin{equation}
    (\sigma^{\prime}, \textbf{c}^{\prime}) = (\sum_{i=1}^{N} \sigma_i \; {mask}_i, \sum_{i=1}^{N} c_i \; {mask}_i)
\end{equation}
where $mask_i$ denotes the membership of the current point to the $i^{\text{th}}$ object.
Additionally, blending weights for dynamic objects must be recalculated:
\begin{equation}
    b^{\prime} = \sum_{i=1}^{N} b_i \; {mask}_i
\end{equation}
In Fig.~\ref{fig:composition+reversal}\textcolor{red}{a}, we create a novel scene by separately using moving objects from the \textit{Balloon2} dataset and backgrounds from the \textit{Playground} dataset, and then compositing them together.
Subsequently, the rendering formula can be employed to compute the pixel color.
The computational complexity of the combination increases in proportion to the number of objects.

\end{document}